\documentclass{article}
\usepackage{PRIMEarxiv}
\usepackage[utf8]{inputenc} % allow utf-8 input
\usepackage[T1]{fontenc}    % use 8-bit T1 fonts
\usepackage{hyperref}       % hyperlinks
\usepackage{url}            % simple URL typesetting
\usepackage{booktabs}       % professional-quality tables
\usepackage{amsfonts}       % blackboard math symbols
\usepackage{nicefrac}       % compact symbols for 1/2, etc.
\usepackage{microtype}      % microtypography
\usepackage{lipsum}
\usepackage{fancyhdr}       % header
\usepackage{graphicx}       % graphics
\graphicspath{{media/}}     % organize your images and other figures under media/ folder

\usepackage{amsmath}
\usepackage{float}
\usepackage{wrapfig}
\usepackage{graphicx}
\usepackage{listings}
\usepackage[frozencache,cachedir=minted-cache]{minted}
\usepackage{amsfonts}
\usepackage{gensymb}
\usepackage{blindtext}
\usepackage{mathtools}
\usepackage{amsthm}
\usepackage{bm}

\newtheorem{theorem}{Theorem}[section]
\newtheorem{lemma}[theorem]{Lemma}
\newtheorem{corollary}{Corollary}[theorem]
\newtheorem{defn}[theorem]{Definition}

\usepackage[toc,page]{appendix}

\title{Mathematical Foundation and Corrections for Full-Range Head Pose Estimation
}

\author{
Hu, Huei-Chung\\
  Docomo Innovations \\
  \texttt{heidi.hu@docomoinnovations.com} \\
    \and
Wu, Xuyang  \\
    Santa Clara University \\
    \texttt{xwu5@scu.edu}
    \and
Wang, Yuan \\
    Santa Clara University \\
    \texttt{ywang4@scu.edu}
    \and
Fang, Yi\thanks{Corresponding authors.} \\ % corresponding author
    Santa Clara University \\
    \texttt{yfang@scu.edu}
    \and
Wu, Hsin-Tai\footnotemark[1]\\ % corresponding author
    Docomo Innovations \\
    \texttt{hwu@docomoinnovations.com}
}

\begin{document}
\maketitle

\begin{abstract}
Numerous works concerning head pose estimation (HPE) offer algorithms or proposed neural network-based approaches for extracting Euler angles 
from either facial key points or directly from images of the head region. However, many works failed to provide clear definitions of the coordinate systems and Euler or Tait–Bryan angles orders in use.
It is a well-known fact that rotation matrices depend on coordinate systems, and \textbf{yaw}, \textbf{roll}, and \textbf{pitch} angles are sensitive to their application order.
Without precise definitions, it becomes challenging to validate the correctness of the output head pose and drawing routines employed in prior works.
In this paper, we thoroughly examined the Euler angles defined in the 300W-LP dataset, head pose estimation such as 3DDFA{\_}v2, 6D-RepNet, WHENet, etc, and the validity of their drawing routines of the Euler angles. When necessary, we infer their coordinate system and sequence of \textbf{yaw}, \textbf{roll}, \textbf{pitch} from provided code. This paper presents
(1) code and algorithms for inferring coordinate system from provided source code, code for Euler angle application order and extracting precise rotation matrices and the Euler angles, (2) code and algorithms for converting poses from one rotation
system to another, (3) novel formulae for 2D augmentations of the rotation matrices, and (4) derivations and code for the correct drawing routines
for rotation matrices and poses. This paper also addresses the feasibility of defining rotations with Wikipedia \& SciPy’s right-handed coordinate system, which makes the Euler angle extraction much easier for full-range head pose research. Our code will be released in \href{https://github.com/foresee-ai/hpe_math}{https://github.com/foresee-ai/hpe{\_}math}.
\end{abstract}

% keywords can be removed

\section{Introduction}
Head Orientation Estimation has evolved significantly, driven by the increasing integration of computer vision and vision-based deep learning. Early approaches often relied on facial key-point detection and images with fixed rotation angle labeling. However, this field has witnessed a paradigm shift with the advent of deep learning, especially convolutional neural networks (CNNs). The ability of deep learning models to automatically learn hierarchical features from data has significantly improved the accuracy and robustness of head orientation estimation algorithms. 

Nevertheless, it is noteworthy that many current works on head pose estimation lack fundamental definitions of the 
coordinate systems employed. Additionally, some methodologies borrowed facial landmark-based code intended for 
limited (mostly camera-facing) head pose estimation and applied to a broader range of head poses. 
Moreover, facial orientation drawing routines are often borrowed from code designed for a different coordinate system. 
This deficit in mathematical clarity also deters the application of standard geometric 2D image augmentations, 
such as flipping and rotation. This paper aims to address these gaps in understanding.

The structure of this paper is as follows: we start by presenting classical Head Pose Estimation (HPE) approaches 
and formulate the foundational mathematical concepts involved. 
Subsequently, we conduct a comprehensive analysis of select datasets, employing symbolic computation techniques where necessary, to infer key parameters, 
particularly when clear definitions of coordinate systems and Euler angles are absent throughout the literature. 
We also delve into the conversion of Euler angles between different coordinate systems, providing code implementations where relevant. 
Furthermore, we identify and highlight inaccuracies or imprecision in both code and literature of some previous works.

Lastly, we present a comprehensive derivation of the transformation of rotation matrices under typical 2D image augmentations. These derivations enable improved model performance without the necessity of increasing the volume of training images. To our knowledge, these derivations have not been utilized or documented in current literature related to Head Pose Estimation (HPE).

\section{Related Work}
This paper will focus on monocular RGB images.
Classical approaches, such as template matching and detector arrays,  related to head pose estimation (HPE) preceding 2008 can be found in the survey paper \cite{hpe_article}.  Deformable models are thoroughly discussed in \cite{1227983}, \cite{DBLP:conf/cvpr/ZhuLLSL16}, \cite{guo2020towards}, and \cite{zhu2017face} and were used to create the commonly used synthetic pose dataset such as 300W{\_}LP \cite{DBLP:conf/cvpr/ZhuLLSL16}.

\subsection{Head pose estimation} 
\begin{figure}[H]
\centering
\includegraphics[width=0.3\textwidth]{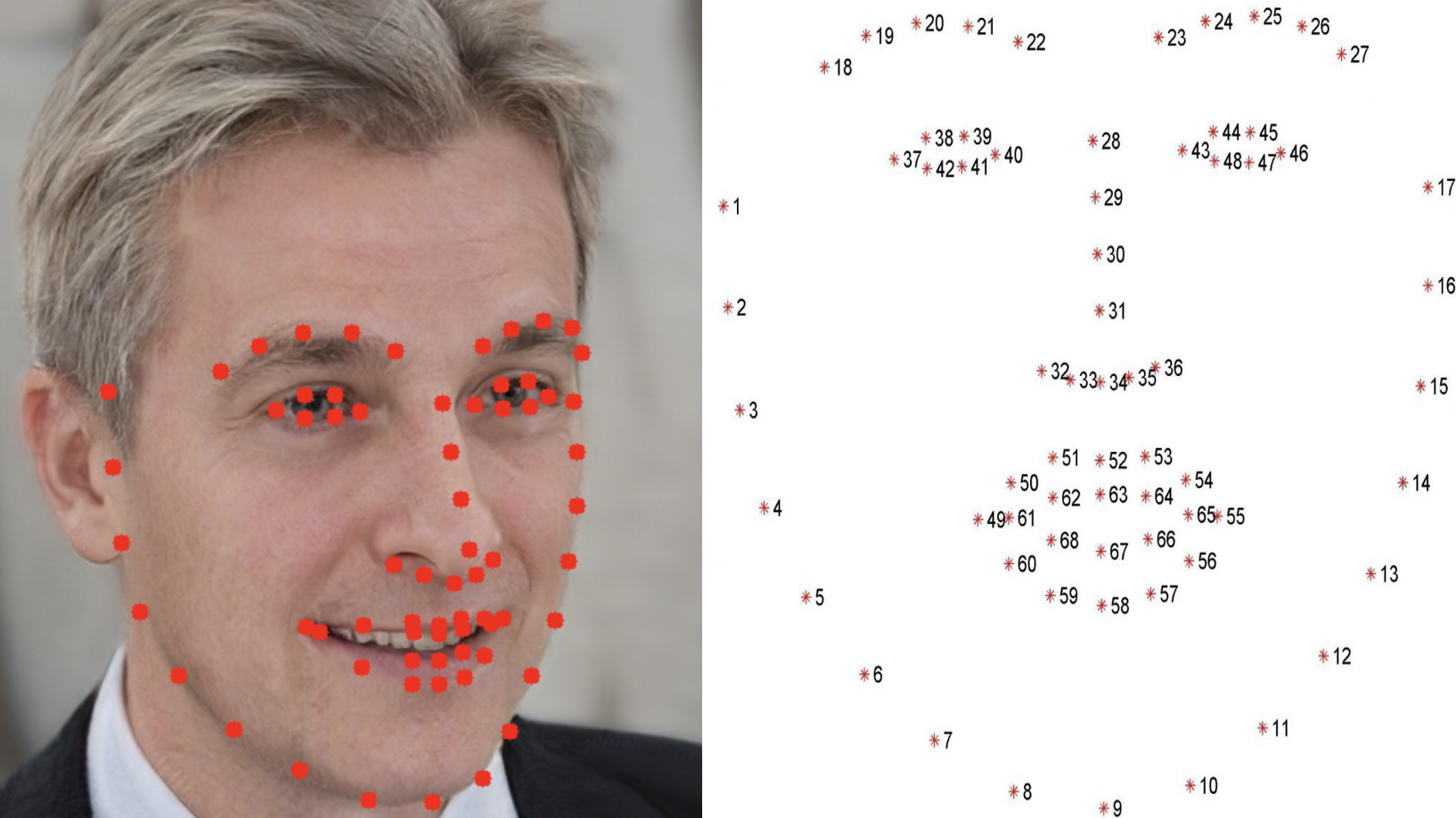}
\caption{Dlib's 68 facial landmarks. \cite{DBLP:journals/vc/ElmahmudiU21} }
\label{fig:dlib}
\end{figure}

Head pose estimation can be divided into two categories, with and without facial landmarks. 
Traditionally, head orientation only makes sense when facial features are visible. See Fig. \ref{fig:dlib} for reference. Obviously, when people are facing away from the camera, no key points can be detected, and this method will not work. Dlib\cite{dlib09} is one of the pioneers in using face landmarks for prediction. 

From facial landmarks, we can mathematically derive the 3D transformations involved in changing the 2D projected shape of facial landmarks with the help of a "standard" face. Later, with the advent of deep learning, researchers start to explore the possibility of utilizing convolutional neural networks to predict the three Euler angles directly, for example, Hopenet\cite{Ruiz_2018_CVPR_Workshops} and WHENet\cite{zhou2020whenet}.  However, the Euler-angle loss easily leads to discontinuity because every angle has multiple numerical representations, e.g., $0\degree = 360\degree$.  To avoid such discontinuity, 6D-RepNet \cite{Hempel_2022} (including 6D-RepNet360\cite{hempel2023robust}) and TriNet\cite{cao2020vectorbased} predict the $3 \times 2$ and $3 \times 3$ rotation matrices respectively. 6D-RepNet still applies the Euler-angle evaluation metric for comparison with other HPE methods. However, due to the natural scarcity of the labeled data, most of these approaches still utilize datasets created/ synthesized from facial key-points approaches. Next, we will review some of the mathematical basics involved in HPE. 

\subsection{Head pose dataset creation}
The 300W across Large Poses (300W-LP) dataset \cite{DBLP:conf/cvpr/ZhuLLSL16} was based on 300W \cite{DBLP:conf/iccvw/SagonasTZP13}, which standardized multiple alignment datasets with 68 landmarks, including AFW \cite{DBLP:conf/cvpr/ZhuR12}, LFPW \cite{DBLP:conf/cvpr/BelhumeurJKK11}, HELEN \cite{DBLP:conf/iccvw/ZhouFCJY13}, IBUG \cite{DBLP:conf/iccvw/SagonasTZP13}, and XM2VTS \cite{Messer1999XM2VTSDBTE}. 
300W{\_}LP's labeled Euler angles are extracted from the rotation matrix estimated from \cite{DBLP:conf/cvpr/ZhuLLSL16}'s 3D Dense Face Alignment (3DDFA), which applies Blanz et al.'s morphable model 3DMM \cite{1227983}, to obtain the standard and rotated 3D faces tailored for an image. While the labeled poses of 300W{\_}LP have been widely used for head pose training or testing, we found multiple 300W{\_}LP's Euler-angle formulas/implementations are used for papers on HPE and 3D dense face alignment [\cite{DBLP:conf/cvpr/ZhuLLSL16}, \cite{3ddfa_cleardusk}, \cite{guo2020towards}, \cite{Hempel_2022}, \cite{hempel2023robust}]. This motivated us to find out which formulas align to 300W{\_}LP's rotation and pose definitions.  

While 300W{\_}LP offered pose labels limited to frontal views, the CMU Panoptic Dataset \cite{Joo_2017_TPAMI} provided extensive 3D facial landmarks for multiple individuals captured from cameras spanning an entire hemisphere, which can be turned into head pose. WHENet \cite{zhou2020whenet} pioneered techniques to transform the CMU Panoptic Dataset into a comprehensive head pose estimation (HPE) dataset covering a larger range of angles. 6D-RepNet360 \cite{hempel2023robust} used WHENet's CMU Panoptic pose data generation method/code to improve its previous range limited 6D-RepNet \cite{Hempel_2022}.  

\section{Rotations and the Euler angles}
\begin{defn} \label{def:rotation_matrix}
A 3D \textbf{rotation matrix} in a 3D Cartesian coordinate system is a transformation matrix $R \in SO(3)$, where $SO(3)$ is the group of invertible 3 $\times$ 3 matrices such that $det(R) = 1$ and $R \times R^{T} = R^{T} \times R = I_{3}$.  Each column of R represents each rotated axis so that R can be used to rotate objects. Then for every (row) vector $\bm{v} \in \mathbb{R}^{3}$, We assume the application of the rotation is to the left of the vector, i.e. the rotated vector $\bm{v}_{rotated}$ is defined as follows: $\bm{v}_{rotated} = (R \times \bm{v}^{T})^{T}$. 
\end{defn}

\begin{defn} \label{def:extrinsic_def}
Here, we assume the existence of two coordinate systems, one with the origin at the center of the head and moving/rotating along with the head, is called \textbf{intrinsic coordinate system}. The other, without loss of generality, coincides with the intrinsic coordinate system initially but does not move along with the head and is called \textbf{extrinsic coordinate system}.  The 3D rotation R, defined by Definition \ref{def:rotation_matrix}, can also be refined by 2 conventions, \textbf{extrinsic} and \textbf{intrinsic} rotations.  \textbf{Intrinsic rotations} \cite{wiki_euler_angle_2023} are elemental rotations that occur about the axes of the intrinsic coordinate system, which axes are denoted in uppercase \textbf{XYZ}. \textbf{Extrinsic rotations} \cite{wiki_euler_angle_2023} are elemental rotations that occur about the axes of the extrinsic coordinate system whose axes are denoted in lowercase, say $xyz$. The \textbf{intrinsic XYZ-coordinate system} remains fixed from the perspective of a targeted human head. In contrast, the \textbf{extrinsic xyz-coordinate system} remains fixed from the perspective of external observers.  

It’s also possible to parameterize the rotation matrix with a smaller set of numbers\cite{wiki_rotation_matrix_2024}.  One of the most common sets of parameters is the \textbf{Euler angles} defined by Leonhard Euler\cite{bhlitem38593}. The Euler angles are the three elemental angles \textbf{yaw, pitch, and roll} representing the orientation of a rigid body to a fixed coordinate system. We use \textbf{handedness}(the right-handed or left-handed rule) along each axis to define the positive direction of each Euler angle. Please note that this is different from using handedness to define left- or right-handed coordinate systems as in some other conventions.  Although they are associated with the three coordinate axes, users must define the desired order to recover the rotation matrices uniquely. 

To avoid ambiguity, we define $\bm{yaw=0, pitch=0},$ and $\bm{roll=0}$ to represent the straight frontal face as shown on the left in Fig. \ref{fig:300w-lp-system} and restrict $\bm{(-\pi, \pi]}$ to be \textbf{the range of yaw, roll, and pitch}. The range of yaw, roll, and pitch is especially important for the loss functions for deep learning based approaches based on the Euler angles.  For example, 300W{\_}LP adapts our definition and its ranges of pitch, yaw, and roll are $(-167\degree, 107\degree), [-90\degree, 90\degree],$ and $(-159\degree, 94\degree)$ respectively. 

To find the Euler angles of a rotation, E. Bernardes's \textit{Quaternion to Euler angles conversion} \cite{Bernardes_2022_pone} describes the mathematical form of decomposing the 3D rotation matrix into the product of three elemental (axis-wise) rotations:
\begin{equation}
    \label{formula:40}
    R = R_{\bm{e_{3}}}(\theta_{3}) \times R_{\bm{e_{2}}}(\theta_{2}) \times R_{\bm{e_{1}}}(\theta_{1})
\end{equation}
where $R_{\bm{e_{i}}}(\theta)$ represents a rotation by the angle $\theta$ around the unit axis vector $\bm{e_{i}}$. These consecutive axes must be orthogonal, i.e., $\bm{e_{1}} \cdot \bm{e_{2}} = \bm{e_{2}} \cdot \bm{e_{3}} = 0$.  In addition, $\bm{e_{1}}$, $\bm{e_{2}}$ and $\bm{e_{3}}$ are orthogonal unit vectors and $\bm{e_{3}} = \epsilon \cdot (\bm{e_{1}} \times \bm{e_{2}})$ in which $\epsilon = (\bm{e_{1}} \times \bm{e_{2}}) \cdot \bm{e_{3}} = \pm1$.

% To summarize, we call a \textbf{properly defined rotation system for HPE} as composed of the following components: (1) A 3D Cartesian coordinate system, (2) Corresponding rotation matrices or yaw/pitch/roll associated with coordinate axes, (3) Handedness and range of the rotations for yaw/pitch/roll, and (4) The \textbf{axis-sequence}.
\end{defn}

The following Lemma demonstrates the difference between the two rotation types.

\begin{lemma} \label{thm:diff_rots}
Suppose $\bm{e_{1}}$ and $\bm{e_{2}}$ are the orthogonal unit axis vectors defined in Definition \ref{def:extrinsic_def}. Given the elemental rotation sequence, first rotating $\theta_{1}$ along $\bm{e_{1}}$, then rotating $\theta_{2}$ along $\bm{e_{2}}$, we can derive that the intrinsic rotation is $\Lambda = R_{\bm{e_{1}}}(\theta_{1}) \times R_{\bm{e_{2}}}(\theta_{2})$, and the extrinsic rotation is $\Sigma= R_{\bm{e_{2}}}(\theta_{2}) \times R_{\bm{e_{1}}}(\theta_{1})$. 
\end{lemma}
\begin{proof}
Note that the extrinsic coordinate system stays the same under rotations and also for angle $\phi$ and axis $\bm{e}$, $R^{-1}_{\bm{e}}(\phi) = R^{t}_{\bm{e}}(\phi) = R_{\bm{e}}(-\phi)$. Observe the extrinsic rotation matrix is just successively applying rotation matrices and trivially $\Sigma = R_{\bm{e_{2}}}(\theta_{2}) \times R_{\bm{e_{1}}}(\theta_{1})$. The intrinsic rotation, denoted by $\Lambda$, can be defined by $\bm{v_{rotated}}^{T} = \Lambda \times \bm{v}^{T} $, where both $\bm{v_{rotated}}$ and $\bm{v}$ are row vectors with the coordinates in the extrinsic $xyz$-coordinate system.  Note from the perspective of the human head, rotating $\phi$ about an axis $\bm{e}$ in the intrinsic coordinate system is equivalent to rotating $-\phi$ extrinsically with intrinsic coordinate system unchanged,  so we have $\bm{v}^{T}  = R_{\bm{e_{2}}}(-\theta_{2}) \times R_{\bm{e_{1}}}(-\theta_{1}) \times \bm{v_{rotated}}^{T}$, $ \implies $ $\bm{v}^{T}  = R^{-1}_{\bm{e_{2}}}(\theta_{2}) \times R^{-1}_{\bm{e_{1}}}(\theta_{1}) \times \bm{v_{rotated}}^{T}$ (especially  for $\theta_{1}  = \theta_{2} =90\degree$, $\bm{v}^{T}  = R^{-1}_{X}(90\degree) \times R^{-1}_{Y}(90\degree) \times \bm{v_{rotated}}^{T}$ as demonstrated in the following figure \ref{fig:ex1}). Then $\Lambda^{-1}  = R^{-1}_{\bm{e_{2}}}(\theta_{2}) \times R^{-1}_{\bm{e_{1}}}(\theta_{1}) = (R_{\bm{e_{1}}}(\theta_{1}) \times R_{\bm{e_{2}}}(\theta_{2}))^{-1}$. So we get $\Lambda = R_{\bm{e_{1}}}(\theta_{1}) \times R_{\bm{e_{2}}}(\theta_{2})$. 
\end{proof}
\begin{figure}[H]
\centering
\includegraphics[width=0.6\textwidth]{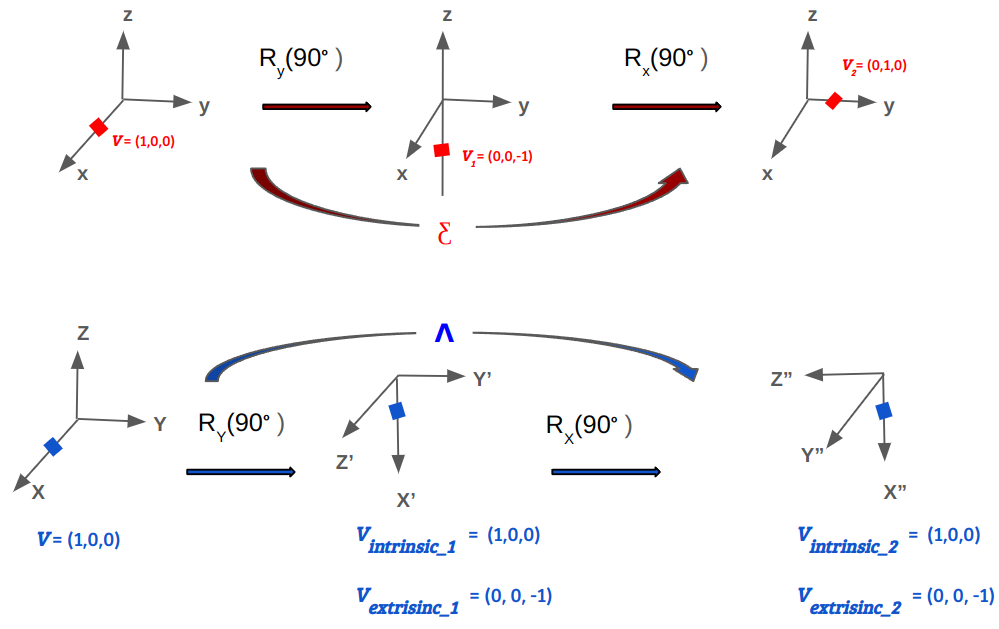}
\caption{Demonstration of successive application of rotations in the case of $\theta_{1} = \theta_2 = 90\degree$}
\label{fig:ex1}
\end{figure} 
\begin{theorem} \label{thm:diff_rots2}
Suppose $\bm{e_{1}}$, $\bm{e_{2}}$, and $\bm{e_{3}}$ are the three orthogonal unit axis vectors defined in Definition \ref{def:extrinsic_def}. Given the elemental rotation sequence, first rotating $\theta_{1}$ along $\bm{e_{1}}$, then rotating $\theta_{2}$ along $\bm{e_{2}}$, last rotating $\theta_{3}$ along $\bm{e_{3}}$,  the intrinsic and extrinsic rotations, denoted by $R_{intrinisc}$ and $R_{extrinisc}$ respectively, can be expressed as follows: 
\begin{equation}
\label{formula:53}
\begin{split}
R_{intrinsic} &= R_{\bm{e_{1}}}(\theta_{1}) \times R_{\bm{e_{2}}}(\theta_{2}) \times R_{\bm{e_{3}}}(\theta_{3}) \\
R_{extrinsic} &= R_{\bm{e_{3}}}(\theta_{3}) \times R_{\bm{e_{2}}}(\theta_{2}) \times R_{\bm{e_{1}}}(\theta_{1}) 
\end{split}
\end{equation}
\end{theorem}
\begin{proof}
Apply Lemma \ref{thm:diff_rots} twice and prove the equality \ref{formula:53} holds.
\end{proof}
Based on Theorem \ref{thm:diff_rots2} and the fact that matrix multiplications are non-commutative, we can conclude (1) the intrinsic and extrinsic rotation matrices are usually different; and (2) \textbf{\emph{the axis-sequence} (the rotating sequence of the three axes) matters} for identifying the multiplicative order of the elemental rotation matrices to compute the real rotation matrix $R$. For example, the intrinsic XYZ-sequence rotation yields   $R_{X} \times R_{Y} \times R_{Z}$ while the intrinsic ZYX-sequence rotation yields $R_{Z} \times R_{Y} \times R_{X}$.

One important observation from Lemma \ref{thm:diff_rots} is every sequence of intrinsic rotations has an equivalent sequence of extrinsic rotations in exact reverse order. 

To summarize, we suggest HPE paper authors or dataset creators give \textbf{\emph{well-defined rotation system for HPE}} which should composed of the following: (1) Specify the 3D Cartesian coordinate system used, (2) Each axis's handedness and associated Euler angle (if applicable), (3) Range of yaw, pitch, and roll (if applicable), and (4) The yaw, pitch, and roll rotation sequence (if applicable).  For simplicity, we also refer a coordinate (or rotation) system A adapting B's system to A's system follows B's system disregarding transitions and scaling.

\section{Wikipedia's intrinsic ZYX-sequence Rotations }
The intrinsic rotation system employed by Wikipedia deserves attention for its widespread adoption among engineers and incorporation into SciPy's implementation for rotations and Euler angles.
  
\begin{defn} \label{def:wikiZYX}
\textbf{Wikipedia's intrinsic ZYX-sequence Rotations} are defined in Wikipedia’s right-handed coordinate system \cite{wiki_euler_angle_2023}, and follows the intrinsic ZYX-sequence, which has the order that first applies $R_{Z}$, then $R_{Y}$, last $R_{X}$. Within Wikipedia's system, yaw, pitch, and roll are defined as the rotations along the Z-, Y-, and X-axes by the angles yaw, pitch, and roll. We have the elemental rotations and rotation matrix, denoted by $R_{S}$, as follows: 
\begin{equation}
\label{formula:42}
 \begin{split}
    R_{X}(p) &= \begin{bmatrix}
    1 & 0 & 0\\
    0 & cos(p) & -sin(p)\\
    0 & sin(p) & cos(p)
    \end{bmatrix}, 
    R_{Y}(y) = \begin{bmatrix}
    cos(y) & 0 & sin(y)\\
    0 & 1 & 0\\
    -sin(y) & 0 & cos(y)
    \end{bmatrix},
    R_{Z}(r) = \begin{bmatrix}   
    cos(r) & -sin(r) & 0\\
    sin(r) & cos(r) & 0\\
    0 & 0 & 1
    \end{bmatrix} \\ 
    \\
   R_{S} &= R_{Z}(y) \times R_{Y}(p) \times R_{X}(r) \\
    &= \begin{bmatrix}
         cos(y)cos(p) & cos(y)sin(p)sin(r)-sin(y)cos(r) & cos(y)sin(p)cos(r)+sin(y)sin(r) \\
         sin(y)cos(p) & sin(y)sin(p)sin(r)+cos(y)cos(r) & sin(y)sin(p)cos(r)-cos(y)sin(r) \\
         -sin(p) & cos(p)sin(r) & cos(p)cos(r)
     \end{bmatrix}. 
 \end{split}
\end{equation}
\end{defn}
$y, p, r$  in Formula \ref{formula:42} are the Euler angles, yaw, pitch, and roll, respectively.
\begin{figure}[H]
\centering
\includegraphics[width=0.2\textwidth]{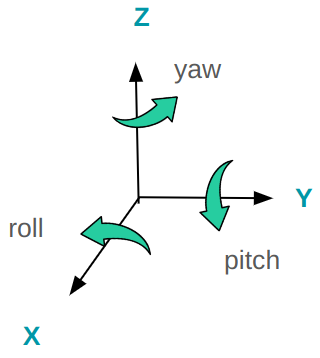}
\caption{SciPy and Wikipedia's coordinate system with right-handed }
\label{fig:scipy_coord}
\end{figure}

%\subsection{SciPy's Euler angle extraction}
According to SciPy Rotation class documentation, \cite{scipy_euler}, its Euler-angle implementation follows Wikipedia's right-handed coordinate system. The convenient script (shown in Listing \ref{listing:9}) allows us to quickly compute the Euler angles (pitch, yaw, and roll) without really dealing with the hassle.   

We re-emphasize here that  Wikipedia and SciPy share the same rotation system, i.e., right-handed intrinsic ZYX-sequence rotations as demonstrated in Fig. \ref{fig:scipy_coord}. 

\subsection{Three-line drawing \& Euler-angle closed-form solution in Wikipedia/SciPy's right-handed coordinate
system} \label{subsec:scipy_3L}
Three-line drawings are generally utilized in HPE papers to visually represent head poses in 2D images. A popular convention is nose direction in blue, neck direction in green, and left face direction in red, as illustrated in Figure \ref{fig:three-line.png}. Under Wikipedia’s coordinate system, for simplicity, we ignore camera intrinsic and regard the three-line drawing as a projection of the coordinate axes $X, Y, -Z$ to the 2D $ZY$ plane, with $X$-axis in blue, $Y$-axis in red and $-Z$-axis in green.

More precisely, before projecting onto the image's 2D Cartesian coordinate system, we first apply a coordinate transformation to the rotation matrix in SciPy's coordinate system into the one with the Z-axis pointing downwards. Then, project the transformed matrix onto the YZ-plane to get the vectors associated with the three RGB lines. As in \ref{formula:43}, the new rotation matrix, $R^{\textbf{draw}}_{S}$, represents the rotation $R_{S}(y, p, r)$ (defined by Formula \ref{formula:42}) under SciPy's coordinate system) in the new 3D coordinate system with Z-axis pointing downward.  Finally, apply the YZ-plane projection.
\begin{equation}
  \label{formula:43}
  \begin{split}
   R^{\bm{draw}}_{S} = T_{S} \times &R_{S}(y, p, r) \times T^{-1}_{S}, \quad where 
   \quad T_{S} = \begin{bmatrix}
                    1 & 0 & 0  \\
                    0 & 1 & 0 \\
                    0 & 0 & -1
                \end{bmatrix}.\\
      and \quad Proj_{YZ} &= \begin{bmatrix}
    0 & 1 & 0 \\
    0 & 0 & 1
    \end{bmatrix}\times \quad R^{\bm{draw}}_{S}. 
  \end{split}
\end{equation}
The above formula, Formula \ref{formula:43}, also reveals one important observation: the key of this drawing (matrix) solely depends on the rotation matrix itself.  Therefore \textbf{the Euler angles are not necessary for the drawing if the rotation matrix is provided}, even if yaw, roll, and pitch are provided they still need to be turned into the associated rotation matrix before the actual drawing.  

%\subsection{Euler-angle closed-form solution for Wikipedia’s right-handed intrinsic ZYX-sequence rotations}
Slabaugh, G. G's \cite{slabaugh1999computing} provides the Euler angle extraction pseudo code for Wikipedia's rotation, and its extracted Euler angles lie within $[0, 2\pi]$. To help visualize the head poses, we prefer the range $(-\pi, \pi]$. So, we modified Slabaugh's pseudo code to suit our needs.  The two pitch-yaw-roll closed-form conversions from a rotation matrix's Python code implementation are shown in Listing (\ref{listing:10}).

Experiments show that the difference between our method (Listing \ref{listing:10}) and SciPy's is minuscule, which is caused by the nature of \href{https://en.wikipedia.org/wiki/Machine_epsilon}{machine epsilon}. SciPy's and our implementations maintain excellent numerical stability; the Frobenius norms of these yaw-roll-pitch solutions only differ from the ground-truth poses roughly within 2 $\times$ machine epsilon. See Fig. \ref{fig:pose_conversion} for the comparison.  

\section{The 300W-LP dataset}
Like 300W dataset \cite{sagonas2016300}, 300W-LP adopts the proposed face profiling to generate 61,225 samples across large poses (1,786 from IBUG, 5,207 from AFW \cite{DBLP:conf/cvpr/ZhuR12}, 16,556 from LFPW \cite{DBLP:conf/cvpr/BelhumeurJKK11}, and 37,676 from HELEN \cite{DBLP:conf/iccvw/ZhouFCJY13}, XM2VTS \cite{Messer1999XM2VTSDBTE} is not used), which is further expanded to 122,450 samples with flipping. The literature itself lacks the basic definition of coordinate and rotation system as mentioned in Section \ref{def:rotation_matrix}, we will try to infer them from the provided code implementations in the following sections.
% heidi found BFM  \hyperlink{https://github.com/jadewu/3D-Human-Face-Reconstruction-with-3DMM-face-model-from-RGB-image/tree/main/BFM}{(BFM)}

\subsection{Derivation of 300W-LP's rotation system  based on its implementation \cite{300w_lp_url} }
\begin{figure}[H]
\centering
\includegraphics[width=0.6\textwidth]{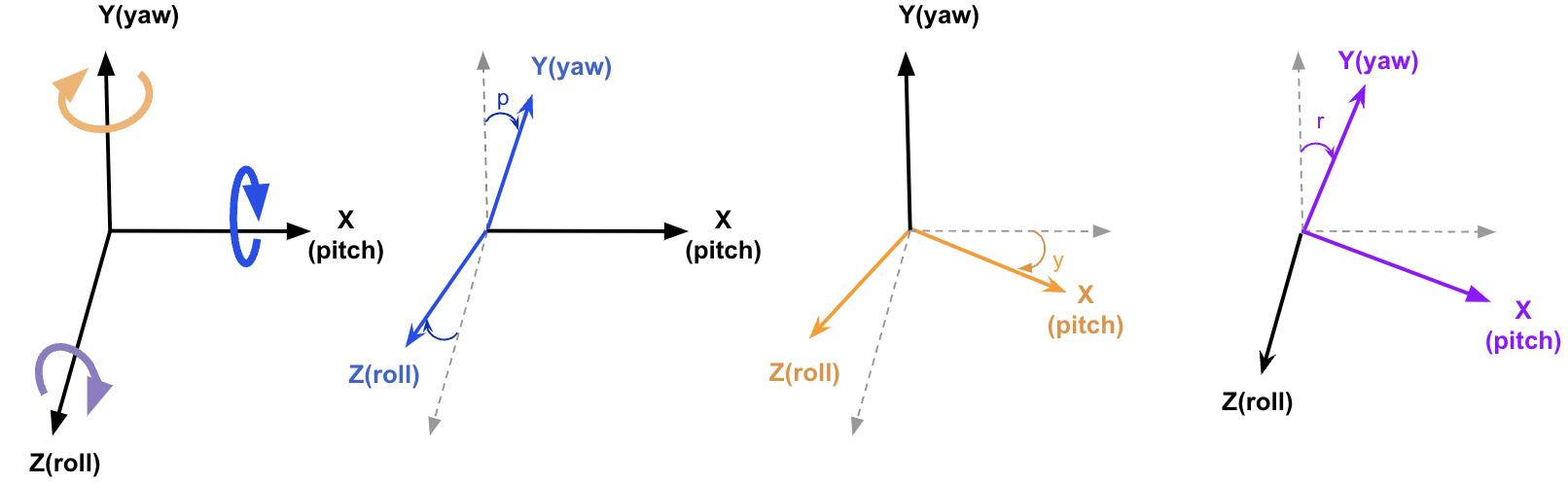}
\caption{300W-LP's elemental rotations}
\label{fig:300w-lp-system2}
\end{figure}

\begin{figure}[H]
\centering
\includegraphics[width=0.6\textwidth]{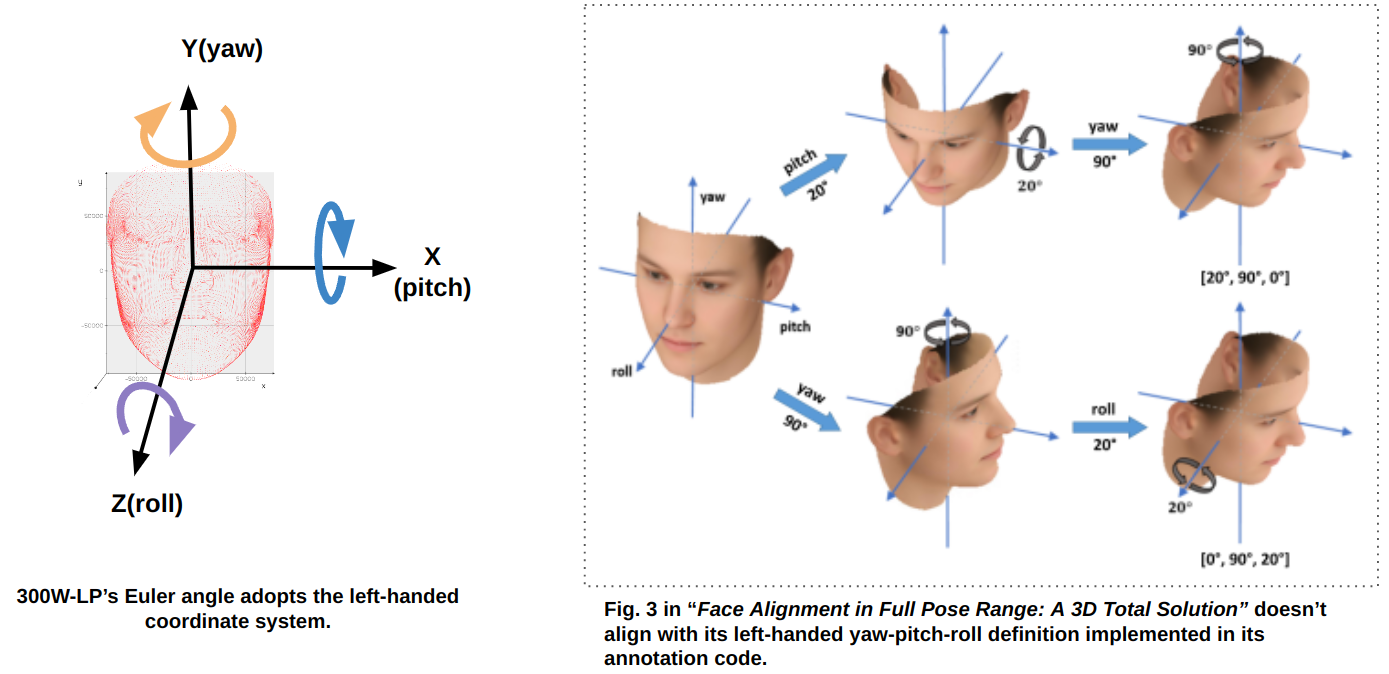}
\caption{300W-LP's Euler angle illustration does not align with what we inferred from its source code}
\label{fig:300w-lp-system}
\end{figure}

While the original code adopts the Base Face Model with neck, its successor \cite{DBLP:conf/cvpr/ZhuLLSL16} uses the \href{https://github.com/cleardusk/3DDFA_V2/issues/101}{BFM} without the ear/neck region. Because both BMFs use the same coordinate system, we decided to use the latter as our Base Face Model (as \cite{300w_lp_url} provided BFM link is no longer valid) to derive 300W{\_}LP's rotation system. First, we can identify the coordinate system used in 300W-LP from the BMF without the neck. Then, the original code provides the elemental rotation matrices for the Euler angles and the axes-sequence (see Formula \ref{eq:16}).  Combining the coordinate system,  elemental rotations, and intrinsic XYZ-sequence (i.e., pitch-yaw-roll order), we conclude that 300W{\_}LP's Euler angles adapt the left-handed rotation. Fig. \ref{fig:300w-lp-system2} demonstrates our inferred definitions for the coordinate system and Euler angle used in 300W-LP's annotations, as shown below. Note $R^{left}_{X}(p)$ means rotation about $X$-axis in left-handed fashion.
\begin{equation*}
    R^{left}_{X}(p) = \begin{bmatrix}
    1 & 0 & 0\\
    0 & cos(p) & sin(p)\\
    0 & -sin(p) & cos(p)
    \end{bmatrix}, 
    R^{left}_{Y}(y) = \begin{bmatrix}
    cos(y) & 0 & -sin(y)\\
    0 & 1 & 0\\
    sin(y) & 0 & cos(y)
    \end{bmatrix}, 
    R^{left}_{Z}(r) = \begin{bmatrix}   
    cos(r) & sin(r) & 0\\
    -sin(r) & cos(r) & 0\\
    0 & 0 & 1
    \end{bmatrix} 
\end{equation*}
\begin{equation}
  \label{eq:16}
  \begin{split}
    R_{W}(y, p, r) &= R^{left}_{X}(p) \times R^{left}_{Y}(y) \times R^{left}_{Z}(r) \\
    & = \begin{bmatrix}
    cos(y)cos(r) & cos(y)sin(r) & -sin(y) \\
    -cos(p)sin(r)+sin(p)sin(y)cos(r) & cos(p)cos(r)+sin(p)sin(y)sin(r) & sin(p)cos(y) \\
    sin(p)sin(r)+cos(p)sin(y)cos(r) & -sin(p)cos(r)+cos(p)sin(y)sin(r) & cos(p)cos(y) 
    \end{bmatrix}
  \end{split}
\end{equation}
However, we found that Fig. 3 on the right (of our Fig. \ref{fig:300w-lp-system} in below) cited from the original 300W-LP paper \cite{DBLP:conf/cvpr/ZhuLLSL16} demonstrates the right-handed Euler angles instead. To avoid further confusion, \textbf{from now on, we stick to the left-handed coordinate system shown in Fig, \ref{fig:300w-lp-system2} and the left in Fig. \ref{fig:300w-lp-system} as 300W-LP's Euler angle definition}.

\subsection{300W-LP's 3D face reconstruction}
\label{sec:300wlp-fp}
\begin{figure}[H]
\centering
\includegraphics[width=0.6\textwidth]{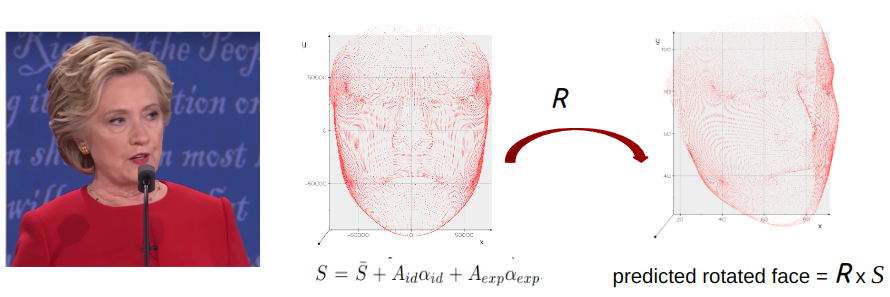}
\caption{Non-scaled, non-transitional, rotated 3D Face by 3DDFA}
\label{fig:bfm2predict}
\end{figure}

\begin{figure}[H]
\centering
\includegraphics[width=0.7\textwidth]{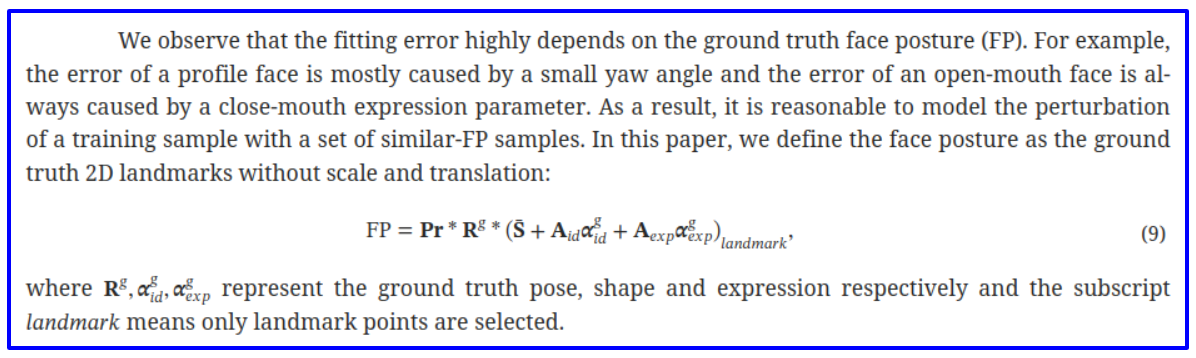}
\caption{The 3DDFA's implementation considers FP perturbation}
\label{fig:3ddfa_why_fp_perturb}
\end{figure}

%We need to first know how to reconstruct the 3D rotated faces from an image to draw a rotation in the images.
The basic idea for obtaining the rotation matrix from an image is to fit the BFM to the human head in the image and then compute the transformation needed to go from the original BFM to the fitted BFM.
\cite{DBLP:conf/cvpr/ZhuLLSL16} proposed the 3D Dense Face Alignment (3DDFA) to obtain the rotated 3D face from an image.  Algorithm 3DDFA applies the 3D morph-able model (3DMM) from Blanz et al.'s \cite{1227983} which describes the 3D face with PCA:
$S = \bar{S} + A_{id}\alpha _{id} + A_{exp}\alpha _{exp}$, where S is a 3D face, $\bar{S}$ is the mean BFM, $A_{id}$ \cite{Paysan2009A3F} is the principle axes trained on the 3D face scans with neutral expression and $\alpha_{id}$ is the shape parameter, $A_{exp}$ \cite{6654137}is the principle axes trained on the offsets between expression scans and neutral scans and $\alpha_{exp}$ is the expression parameter.  The 3D face is then projected onto the image plane with a Weak Perspective Projection:
$V (p) = f \ast Pr \ast R \ast S + t_{2d}$, where $V(p)$ is the 2D positions of model vertexes, $f$ is the scale factor, $Pr = \begin{bmatrix} 
1 & 0 & 0 \\
0 & 1 & 0
\end{bmatrix}$
, $R$ is the rotation matrix calculated from the Euler angles, and $t_{2d}$ is the translation
vector. Fig. \ref{fig:bfm2predict} shows the predicted 3D face model through the non-scaling, non-transition, and rotation-only computation.  
The cited paragraph (shown in Fig. \ref{fig:3ddfa_why_fp_perturb}), from Section 5.1,  in \cite{DBLP:conf/cvpr/ZhuLLSL16} explains how \hypertarget{fpPerturb}{its 3DDFA implementation prevents the fitting errors due to} a profile face rotated by a small yaw angle or the default closed-mouth parameters for an open-mouth face. To avoid such errors, it defines the ground-truth rotation ${R^{g}}$ to be the one from $S = \bar{S} + A_{id}\alpha_{id} + A_{exp}\alpha_{exp}$ to the face points (FP), instead of the one from the BFM (i.e., $\bar{S}$) to FP. We name this approach the \textbf{FP-perturbation} scheme.    

\subsection{300W-LP's drawing method for the Euler angles}
\label{300wlp_euler_drawing}
\begin{figure}[H]
  \begin{minipage}[c]{0.5\linewidth}
  \includegraphics[width=\textwidth,height=3cm]{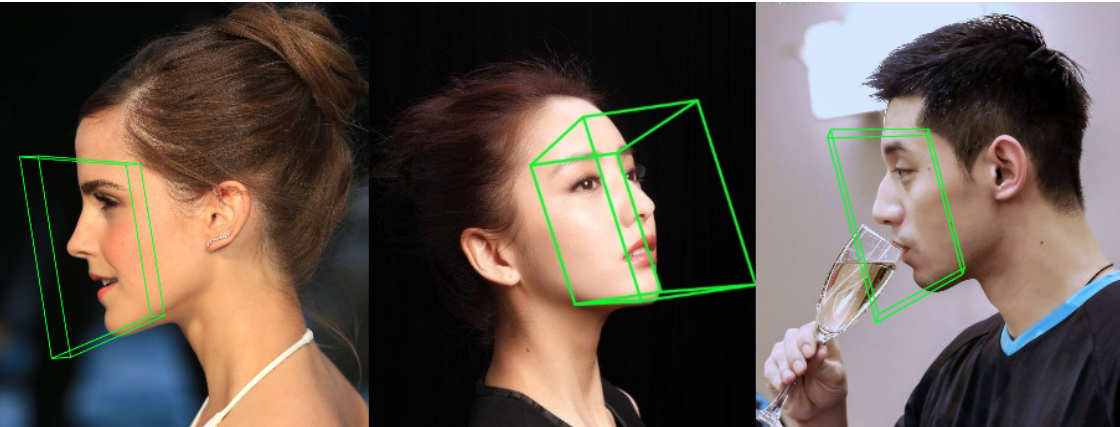}
  \caption{3DDFA's Head Pose Drawing, cited from \cite{3ddfa_cleardusk}}
  \label{fig:pose_box}
  \end{minipage}
  \hfill
  \begin{minipage}[c]{0.5\linewidth}
  \includegraphics[width=\textwidth,height=3cm]{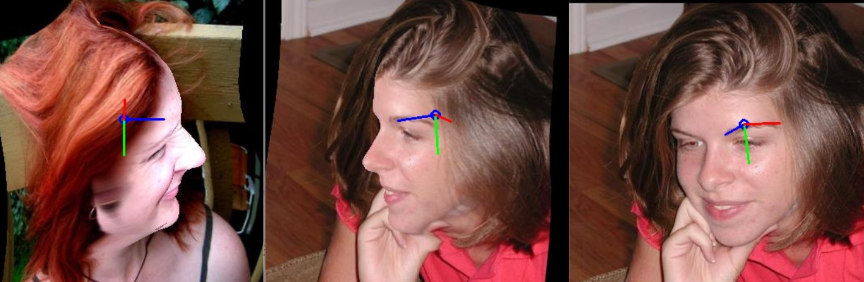}
  \caption{Our Proposed Head Pose Drawing}
  \label{fig:three-line.png}
  \end{minipage}%
% \label{fig:6d}
\end{figure}

Since the code of \cite{DBLP:conf/cvpr/ZhuLLSL16} for 300W{\_}LP dataset didn't provide any drawing methods about rotation, some of the later works, such as 3DDFA and 3DDFA{\_}v2 adopt the scale orthographic projection, in which it reconstructs the camera extrinsic P, consisting of a non-scaled rotation matrix, i.e., $R_{W}$, and 3D offset, and uses the estimated 3D facial vertices to draw the 3D pose frustums.  See Fig. \ref{fig:pose_box}. But while the pose-frustum drawing serves the purpose,  we find it harder to visualize the nuances among rotations, especially when comparing the ground truth and estimated Euler angles.    

In contrast, the three-line drawing, shown in Fig. \ref{fig:three-line.png}, allows us to visualize the minute differences among the rotations.  Here, we follow the convention for drawing, mentioned in section \ref{subsec:scipy_3L}: the red line stretching toward the left side of the frontal face, the green line stretching from the forehead to the center of the jaw, and the blue line pointing from the nose out.  The three lines are perpendicular to each other in the 3-dimensional space, even though sometimes they don't appear so due to 2D projection.  
% \begin{figure}[H]
% \centering
% \includegraphics[width=0.45\textwidth]{figures/pose_box.png}
% \caption{3DDFA's Head Pose Drawing, cited from \cite{3ddfa_cleardusk}}
% \label{fig:pose_box}
% \end{figure}
% \begin{figure}[H]
% \centering
% \includegraphics[width=0.45\textwidth]{figures/three-line.png}
% \caption{300W-LP dataset with the three-line drawing}
% \label{fig:three-line.png}
% \end{figure}

\subsection{Our three-line drawing method for 300W{\_}LP's pose annotations}
\label{our_300wlp_drawing_section}
We can apply a similar procedure in Section \ref{subsec:scipy_3L} by (1) first applying the matrix $T_{W}$ to transform the rotation matrix of 300W{\_}LP's coordinate system into the one of the Y-axis pointing downwards, and (2) then project the transformed matrix onto the XY-plane of the image's system. Now, we derive Formula \ref{formula:30} so that the new matrix, $R^{\textbf{draw}}_{W}$, represents the rotation $R_{W}(y, p, r)$ (defined by Formula \ref{eq:16}) under 300{\_}LP's coordinate system) in the 3D coordinate system associated with the image's system.  Then apply the XY-plane projection.
\begin{equation}
  \label{formula:30}
  R^{\textbf{draw}}_{W} = T_{W} \times R_{W}(y, p, r) \times T^{-1}_{W}, \quad where \quad 
   T_{W} = \begin{bmatrix}
    1 & 0 & 0  \\
    0 & -1 & 0 \\
    0 & 0 & 1
    \end{bmatrix}
\end{equation}

\noindent
The first row of Fig. \ref{fig:3drawings} shows that our proposed three-line drawing routine for 300W{\_}LP's pose annotations works.  Another widely used three-line drawing is the drawing routine \textbf{draw{\_}axis}()  (that has been adopted by many papers as shown in Listing \ref{listing:5}) which we will carefully compare with our drawing routine in section \ref{sec:300wlp_3line_drawing}.

\subsection{Extracting the Euler angles from a rotation matrix in 300W-LP's rotation system}
\label{subsec:extract_300w_euler}
Now that we have identified the rotations utilized in 300W-LP following Formula (\ref{eq:16}), we can proceed to compute the Euler angles from it. We follow Slabaugh, G. G's \cite{slabaugh1999computing} Euler angles' computation except that the range $(-\pi, \pi]$ for the yaw, pitch, and roll values are selected. Although for the pitch and roll, we mainly focus on $[-\pi/2, \pi/2]$ because most head poses occur in this range, including the upside-down human head pose as a result of image augmentations, we cannot eliminate the possibility that true full range may be useful in some occasions. We also use NumPy for trigonometric calculation in code implementations.  

For a given rotation $R=(R_{i,j})$ from 300W-LP where $R_{i,j}$ is the $(i,j)$-the entry in R, we can set up the matrix equation (\ref{eq:17}) to solve for yaw, pitch, and roll.  
\begin{equation}
\label{eq:17}
\begin{split}
(R_{i,j}) &= \begin{bmatrix}
    cos(y)cos(r) & cos(y)sin(r) & -sin(y) \\
    -cos(p)sin(r)+sin(p)sin(y)cos(r) & cos(p)cos(r)+sin(p)sin(y)sin(r) & sin(p)cos(y) \\
    sin(p)sin(r)+cos(p)sin(y)cos(r) & -sin(p)cos(r)+cos(p)sin(y)sin(r) & cos(p)cos(y) 
\end{bmatrix} 
\end{split}
\end{equation}

\bigbreak
Equation $R_{0,2} = -\bm{sin}(y)$ gives yaw's first solution $y_{1} = \bm{arcsin}(-R_{0,2}) \in [-\pi/2, \pi/2]$ due to Numpy's $\bm{arcsin}$ function call.  To restrict both yaws in $(-\pi, \pi]$, we derive the second formula of yaw in (\ref{eq:18}):

\begin{equation}
\label{eq:18}
  y_{2} =
    \begin{cases}
      \pi - y_{1} &  if \verb|  | y_{1} \ge 0  \\
      -\pi - y_{1} & if \verb|  |  y_{1} < 0  
    \end{cases}       
\end{equation}
\noindent
Gimbal locks happen when $\bm{cos}(y) = 0$, i.e, $y_1 = y_2 \text{ and } |y_i| = \pi/2$.  We will discuss it later. 

For the non-Gimbal-lock cases, equivalently $\bm{cos}(y) \neq 0$, we consider the two equations shown in (\ref{eq:19}).
\begin{equation}
\label{eq:19}
    \begin{cases}
      R_{1,2} &= \bm{sin}(p)\bm{cos}(y) \\
      R_{2,2} &= \bm{cos}(p)\bm{cos}(y)
    \end{cases}       
\end{equation}
Adapting to Numpy's $\bm{arctan2}$ makes the solution unique. When applying $\bm{arctan2}$, we must be careful when $\bm{cos}(y) < 0$, as mentioned in \cite{slabaugh1999computing}.  It will result in the sign change for both terms, $\bm{sin}(p)$ and $\bm{cos}(p)$, and lead to the wrong angle prediction, $p \pm \pi$, instead of $p$.  To avoid such a pitfall, we set pitch $p = \textbf{arctan2}(R_{1,2}/\bm{cos}(y), R_{2,2}/\bm{cos}(y))$, which results in $p$ in $[-\pi, \pi]$. From Numpy's $\bm{arctan2}$’s document \cite{arctan}, $p$ can be further restricted into the range $(-\pi, \pi)$ due to both numerator and denominator can’t be $\pm \inf$ and $\pm 0$ at the same time.  Now, we can formulate pitch as follows:
\begin{equation}
\label{eq:20}
    \begin{cases}
      p_{1} &= \bm{arctan2}(R_{1,2}/\bm{cos}(y_{1}), R_{2,2}/\bm{cos}(y_{1}))\\
      p_{2} &= \bm{arctan2}(R_{1,2}/\bm{cos}(y_{2}), R_{2,2}/\bm{cos}(y_{2}))
    \end{cases}       
\end{equation}
\noindent
Formula (\ref{eq:18}) implies that $\bm{cos}(y_2) = -\bm{cos}(y_1)$. Now substitute it back to Formula \ref{eq:20}, then we get $\bm{sin}(p_2) = -\bm{sin}(p_1)$ and $\bm{cos}(p_2) = -\bm{cos}(p_1)$.  Hence $p_{1}$ and $p_{2}$ differs by $\pi$.  To get both of them inside $(-\pi, \pi]$,  we get a simplified formula for pitch, shown in (\ref{align:21-22}):  
\begin{align} 
  \label{align:21-22}
    p_{1} &= \bm{arctan2}(R_{1,2}/\bm{cos}(y_{1}), R_{2,2}/\bm{cos}(y_{1}))\\
    p_{2} &= \begin{cases}
               p_{1} - \pi &\textbf{  if } p_{1} \ge 0 \\
               p_{1} + \pi &\textbf{  if } p_{1} < 0 
            \end{cases}   
\end{align}
\noindent
Similarly, we can derive the formula for roll, as shown in (\ref{align:23-24}):
\begin{align} 
  \label{align:23-24}
    r_{1} &= \bm{arctan2}(R_{0,1}/\bm{cos}(y_{1}), 
              R_{0,0}/\bm{cos}(y_{1}))\\
    r_{2} &= \begin{cases}
               r_{1} - \pi &\textbf{  if } r_{1} \ge 0 \\
               r_{1} + \pi &\textbf{  if } r_{1} < 0 
            \end{cases}   
\end{align}
Gimbal locks occur at $y=\pm \pi/2$. If $y = \pi/2$, $\bm{sin}(y) = 1$ and $\bm{cos}(y)=0$ reduce the matrix (\ref{eq:17}) into (\ref{eq:25}). In which case, we also find that $R_{1,0} = \bm{sin}(p - r)$ and $R_{1,1} = \bm{cos}(p - r)$.  So we have $p-r=\bm{arctan2}(R_{1,0}, R_{1,1})$. While there are infinitely many solutions for $p$ and $r$, to prevent $|p| > \pi/2$, we decide to set $p =  \bm{arctan2}(R_{1,0}, R_{1,1}) / 2 $ and $r = -p$ to force them to fall within $[-\pi/2, \pi/2]$.  
\begin{equation}
\label{eq:25}
\begin{split}
R(p, \pi/2, r) &= \begin{bmatrix}
    0 & 0 & -1 \\
    -cos(p)sin(r)+sin(p)cos(r) & cos(p)cos(r)+sin(p)sin(r) & 0 \\
    sin(p)sin(r)+cos(p)cos(r) & -sin(p)cos(r)+cos(p)sin(r) & 0
    \end{bmatrix} \\ \\
    p &=  \bm{arctan2}(R_{1,0}, R_{1,1}) / 2,  \textbf{   } r = -p
\end{split}
\end{equation}
The other Gimbal lock case is when $y = -\pi/2$. Then the matrix in (\ref{eq:17}) becomes (\ref{eq:26}). Similarly, we can derive $R_{1,0}=-\bm{sin}(p+r)$, $R_{1,1} = \bm{cos}(p+r)$ and $p+r = \bm{arctan2}(-R_{1,0}, R_{1,1})$.  Again, to make both $p$ and $r$ in $[-\pi/2, \pi/2]$, we set $p=r=\bm{arctan2}(-R_{1,0}, R_{1,1}) / 2$.
\begin{equation}
\label{eq:26}
\begin{split}
R(p, -\pi/2, r) &= \begin{bmatrix}
    0 & 0 & 1 \\
    -cos(p)sin(r)-sin(p)cos(r) & cos(p)cos(r)-sin(p)sin(r) & 0 \\
    sin(p)sin(r)-cos(p)cos(r) & -sin(p)cos(r)-cos(p)sin(r) & 0
    \end{bmatrix}, \\ \\
    p =  r &= \bm{arctan2}(-R_{1,0}, R_{1,1}) / 2 \\
\end{split}
\end{equation}
Listing \ref{listing:1} presents the full Python code for computing the closed-form yaw, roll, and pitch for the 300W-LP. 

\begin{figure}[H]
\centering
\includegraphics[width=0.7\textwidth]{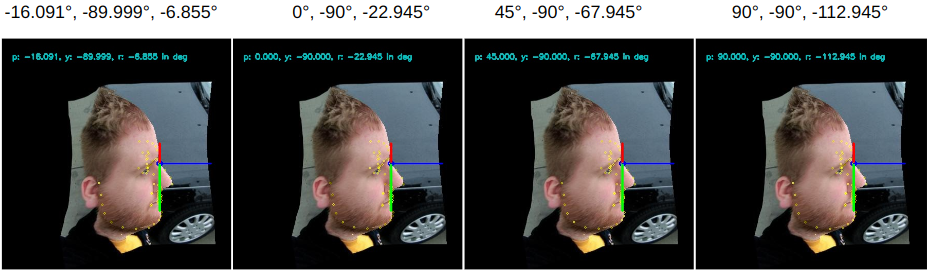}
\caption{300W{\_}LP's Gimbal Lock case.}
\label{fig:gimbal_lock}
\end{figure}

The 300W{\_}LP system also encounters Gimbal locks. Take one of the 300W{\_}LP's image HELEN/HELEN{\_}2375918801{\_}1{\_}14.jpg as an example, its pitch-yaw-roll annotation is (-16.090911401458296\degree, -89.9985818251308\degree, -6.854511900533989\degree) and shown leftmost in Fig. \ref{fig:gimbal_lock}. Even though the labeled yaw $\ne 90\degree$, its corresponding rotation matrix $M$ does meet the Gimbal lock requirement, i.e., $y = -90\degree$ because the Euler angles extracted from $M$ yield the general pitch-yaw-roll solutions, $(p, -90\degree, r)$ and $p+r=-22.94542388660367\degree$. The three images to the right in Fig. \ref{fig:gimbal_lock} show three possible solutions of Euler angles for the same rotation matrix. This indicates that directly learning the Euler angles for full-range HPE is a bad idea unless the ambiguity problem is properly handled.  Other examples are shown in Fig. \ref{fig:gimbol_lock2}. We suggest learning from the rotation matrix representations.  

Furthermore, between the two solutions of yaw, pitch, and roll for the non-gimbal lock case, which one did 300W{\_}LP choose?  According to the dataset's yaw $\in [-90\degree, 90\degree]$, and $y_{1} = \bm{arcsin}(-R_{0,2}) \in [-90\degree, 90\degree]$,  we derive that 300W{\_}LP picked the first solution in Listing \ref{listing:1}.

\section{300W{\_}LP's three-line drawing: Sec \ref{our_300wlp_drawing_section} v.s. draw{\_}axis() (Listing \ref{listing:5})}
\label{sec:300wlp_3line_drawing}
Hopenet \cite{Ruiz_2018_CVPR_Workshops} abandons the traditional approach that entirely relied on landmark detection and then solves the 2D to 3D correspondence problem with a mean human head model. Instead of the extraneous head model and an ad-hoc fitting step, Hopenet determines pose by training its convolutional neural network on the 300W-LP dataset. Its three-line (pose) drawing routine for the 300W{\_}LP dataset is \textbf{draw{\_}axis}() (shown in Listing \ref{listing:5}).  Since then, this drawing routine has been widely used for many head-pose-estimation papers such as WHENet \cite{zhou2020whenet}, 6D-RepNet \cite{Hempel_2022}, DirectMHP \cite{zhou2023directmhp} etc. 

While \textbf{draw{\_}axis}() is popular and seems to depict three lines well, we still need to prove its drawing correct. Due to our earlier discussion in Section \ref{our_300wlp_drawing_section}, we have successfully derived Formula \ref{formula:30}. Let's expand the formula and substitute $R_{W}$ by Formula \ref{eq:16} and $T_{W}$ into $R^{draw}_{W}$. Then we get the following formula:
\begin{equation}
\label{formula:44}
R^{draw}_{W} = \begin{bmatrix}
    cos(y)cos(r) & -cos(y)sin(r) & -sin(y) \\
    cos(p)sin(r)-sin(p)sin(y)cos(r) & cos(p)cos(r)+sin(p)sin(y)sin(r) & -sin(p)cos(y) \\
    sin(p)sin(r)+cos(p)sin(y)cos(r) & sin(p)cos(r)-cos(p)sin(y)sin(r) & cos(p)cos(y) 
\end{bmatrix}. 
\end{equation}
To project the three columns of $R^{draw}_{W}(p, y, r)$ onto the (image's) XY-plane, we can reduce the matrix to the following:
\begin{equation}
\label{formula:45}
R^{draw}_{img}(p, y, r) = \begin{bmatrix}
    cos(y)cos(r) & -cos(y)sin(r) & -sin(y) \\
    cos(p)sin(r)-sin(p)sin(y)cos(r) & cos(p)cos(r)+sin(p)sin(y)sin(r) & -sin(p)cos(y)
\end{bmatrix},
\end{equation}
and the XY-projections for all 3 unit axis vectors are as follows:
\begin{equation}
\label{formula:46}
\begin{split}
\bm{x} &= \begin{bmatrix}
    cos(y)cos(r) \\
    cos(p)sin(r)-sin(p)sin(y)cos(r)
\end{bmatrix}, 
\bm{y} = \begin{bmatrix}
    -cos(y)sin(r) \\
    cos(p)cos(r)+sin(p)sin(y)sin(r) 
\end{bmatrix},
\bm{z} = \begin{bmatrix}
    -sin(y) \\
    -sin(p)cos(y) 
\end{bmatrix}
\end{split}
\end{equation}
To accommodate Line 3's turning yaw to -yaw in \textbf{draw{\_}axis}() (Listing \ref{listing:5}), we replace $cos(y) = cos(-y)$ and $sin(y) = -sin(-y)$ into Formula \ref{formula:46} to get Formula \ref{formula:47}. Next, follow Line 3 set $\tilde{y} = -y$ in Formula \ref{formula:47} and we get Formula \ref{formula:48} 
\begin{equation}
\label{formula:47}
\begin{split}
\bm{x'} &= \begin{bmatrix}
    cos(-y)cos(r) \\
    cos(p)sin(r)+sin(p)sin(-y)cos(r)
\end{bmatrix},
\bm{y'} = \begin{bmatrix}
    -cos(-y)sin(r) \\
    cos(p)cos(r)-sin(p)sin(-y)sin(r) 
\end{bmatrix},
\bm{z'} = \begin{bmatrix}
    sin(-y) \\
    -sin(p)cos(-y)
\end{bmatrix};
\end{split}
\end{equation}
\begin{equation}
\label{formula:48}
\begin{split}
    \begin{bmatrix}
        \textrm{x1} \\ 
        \textrm{y1}
    \end{bmatrix} &= \begin{bmatrix}
    cos(\tilde{y})cos(r) \\
    cos(p)sin(r)+sin(p)sin(\tilde{y})cos(r)
    \end{bmatrix},
    \begin{bmatrix}
        \textrm{x2} \\ 
        \textrm{y2}
    \end{bmatrix}  = \begin{bmatrix}
    -cos(\tilde{y})sin(r) \\
    cos(p)cos(r)-sin(p)sin(\tilde{y})sin(r) 
    \end{bmatrix},
    \begin{bmatrix}
        \textrm{x3} \\ 
        \textrm{y3}
    \end{bmatrix}  = \begin{bmatrix}
    sin(\tilde{y}) \\
    -sin(p)cos(\tilde{y}) 
    \end{bmatrix}.
\end{split}
\end{equation}
Since Formula \ref{formula:48} completely matches with Lines 15-25 of \textbf{draw{\_}axis}() (Listing \ref{listing:5}), we proved  \textbf{draw{\_}axis}() equals to our proposed drawing routine (Formula \ref{formula:46}). 
% \begin{table}[!h]
% \centering
% \begin{tabular}{|c|c|c|c|} 
%  \hline
%  pose & left & middle & right \\ [0.1ex] 
%  \hline
%  pitch & 6.208 & -17.325 & -7.601 \\ [0.1ex] 
%  \hline 
%  yaw & 5.876 & -49.589 & -54.009 \\ [0.1ex] 
%  \hline
%  roll & -1.694 & 11.423 & 4.450 \\ [0.1ex] 
%  \hline
% \end{tabular}
% \caption{300W{\_}LP's pose labels for Fig. \ref{fig:3drawings}}
% \label{table:300wlp_labels}
% \end{table}

% \begin{figure}[H]
% \centering
% \includegraphics[width=0.5 \textwidth]{figures/3drawings.png}
% \caption{Draw Table \ref{table:300wlp_labels}'s poses with our 300W{\_}LP's drawing routine (top) and \textbf{draw{\_}axis} (down). The images come from 300W{\_}LP's HELEN, LFPW, and IBUG datasets. We can see that both drawings produce the same outputs.}
% \label{fig:3drawings}
% \end{figure}
\begin{table*}
    \begin{minipage}{0.4\linewidth}
        \label{table:300wlp_labels}
        \centering
\begin{tabular}{|c|c|c|c|} 
 \hline
 pose & left & middle & right \\ [0.1ex] 
 \hline
 pitch & 6.208 & -17.325 & -7.601 \\ [0.1ex] 
 \hline 
 yaw & 5.876 & -49.589 & -54.009 \\ [0.1ex] 
 \hline
 roll & -1.694 & 11.423 & 4.450 \\ [0.1ex] 
 \hline
\end{tabular}
     \caption{300W{\_}LP's pose labels for Fig. \ref{fig:3drawings}}
    \end{minipage}
    \begin{minipage}{0.5\linewidth}
    \begin{figure}[H]
        \centering
            \includegraphics[width=.9\textwidth]{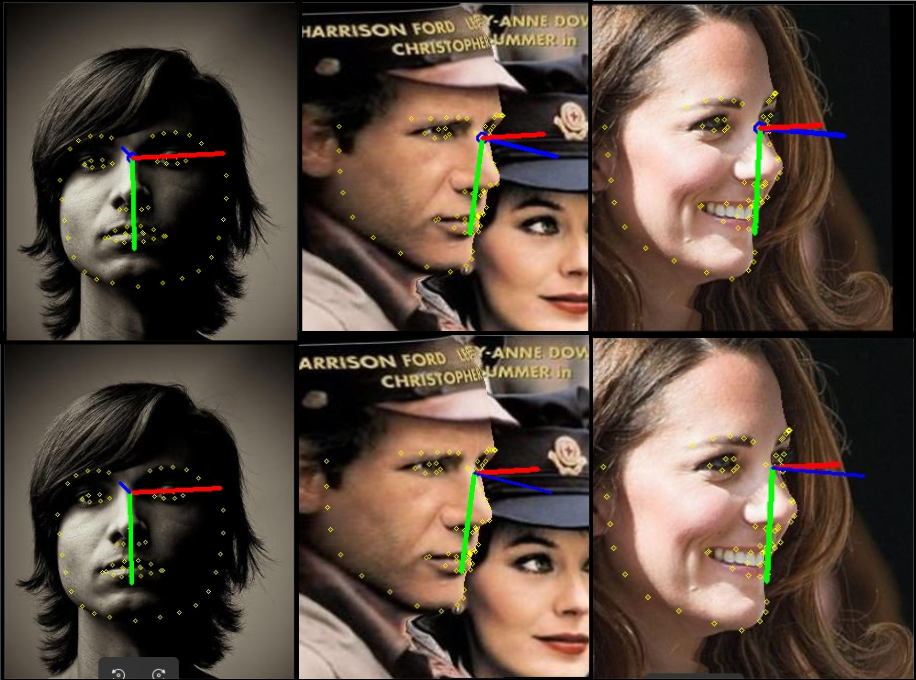}
    \caption{Draw Table 1's poses with our 300W{\_}LP's drawing routine (top) and \textbf{draw{\_}axis} (down). Images come from 300W{\_}LP's HELEN, LFPW, and IBUG datasets. Both methods produce the same outputs.}
    \label{fig:3drawings}
    \end{figure}
    \end{minipage}
\end{table*}
\section{Euler angle Conversion between SciPy's intrinsic ZYX-order and 300W{\_}LP's systems}
\label{sec:pose_conversion}
In scenarios such as creating new datasets, adhering to a consistent coordinate system is imperative. A prevalent practice is to employ a popular one such as the 300W{\_}LP or SciPy's coordinate/rotation system. Consequently, the necessity of performing conversions of Euler angles between these coordinate systems may arise.
\begin{figure}[H]
\centering
\includegraphics[width=0.5\textwidth]{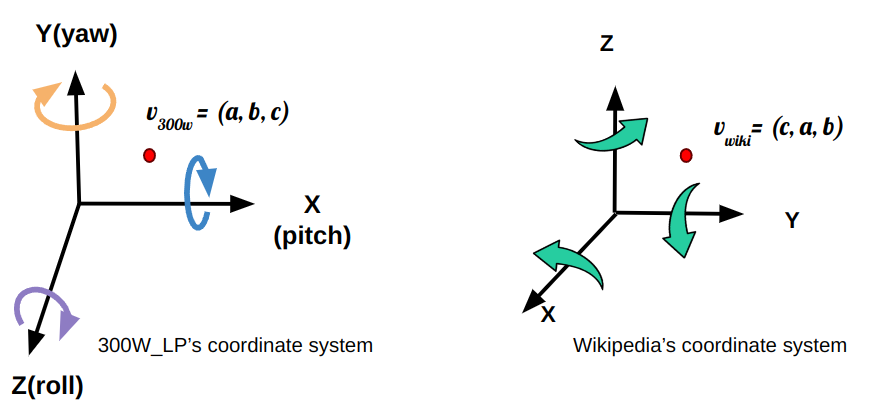}
\caption{300W{\_}LP and Wikipedia's coordinate systems.}
\label{fig:coord_sys}
\end{figure}

\subsection{Convert 300W{\_}LP's rotations into SciPy's}
To convert a 300W{\_}LP's rotation $R_{W}$ into SciPy's $R_{S}$, we first substitute a given triple of yaw, pitch, and roll into Formula \ref{eq:16} to obtain $R_{W}$; next apply a coordinate transformation to transform $R_{W}$ of 300W{\_}LP’s coordinate system into its converted rotation matrix, denoted by $\mathbb{T}_{w2s}$, of SciPy's; and finally compute SciPy's intrinsic ZYX-order Euler angles through plugging $R_{W}$ into extract{\_}wikiZYX{\_}pose() in Listing \ref{listing:10}.  
Fig. \ref{fig:coord_sys} shows the relation between 300W{\_}LP and  SciPy's coordinate systems. So the matrix 
\begin{equation}
\label{formula:56}
T_{w2s} \coloneqq \begin{bmatrix}
    0 & 0 & 1 \\
    1 & 0 & 0\\
    0 & 1 & 0    
\end{bmatrix}
\end{equation} 
will transform a vector, say $\bm{v}_{300W} = (a, b, c)$, in 300W{\_}LP's coordinate system into one defined in SciPy's coordinate system, $\bm{v}_{wiki} = (c, b, a)$. Then $\mathbb{T}_{w2s}$ mentioned earlier can be defined by:
\begin{equation}
\label{formula:54}
\begin{split}
\mathbb{T}_{w2s} = T_{w2s} \times R_{W} \times T^{-1}_{w2s}
    &= \begin{bmatrix}
    0 & 0 & 1 \\
    1 & 0 & 0\\
    0 & 1 & 0    
   \end{bmatrix}  \times R_{W} \times \begin{bmatrix}
    0 & 0 & 1 \\
    1 & 0 & 0\\
    0 & 1 & 0    
    \end{bmatrix} \\
   &= \begin{bmatrix}
    0 & 1 & 0 \\
    0 & 0 & 1\\
    1 & 0 & 0     
   \end{bmatrix}  \times R_{Z}(y) \times R_{Y}(p) \times R_{X}(r) \times \begin{bmatrix}
    0 & 1 & 0 \\
    0 & 0 & 1\\
    1 & 0 & 0    
    \end{bmatrix} ,
\end{split}
\end{equation}
where $T_{w2s}$ and $y, p, r, R_{Z}, R_{Y}, R_{X}$ are defined in Formula \ref{formula:56} and \ref{formula:42} respectively.
\subsection{Convert SciPy's rotations into 300W{\_}LP's}
Similarly, to convert a SciPy's rotation $R_{S}$ into 300W{\_}LP's $R_{W}$, we first substitute a given triple of yaw, pitch, and roll into Formula \ref{formula:42} to obtain $R_{S}$; next apply a coordinate transformation, denoted by $\mathbb{T}_{s2w}$ to transform $R_{S}$ of SciPy's coordinate system into $R_{W}$ of 300W{\_}LP's; and finally compute 300W{\_}LP's intrinsic XYZ-order Euler angles through plugging $R_{W}$ into extract{\_}angles{\_}for{\_}300W() in Listing \ref{listing:1}.  
The matrix $T_{s2w} \coloneqq T^{-1}_{w2s}$ will transform a vector, say $\bm{v}_{wiki} = (c, b, a)$, in SciPy's coordinate system into the vector $\bm{v}_{300W} = (a, b, c)$ in 300W{\_}LP's coordinate system, Then $\mathbb{T}_{s2w}$ in (2) is defined by:
\begin{equation}
\label{formula:55}
\begin{split}
\mathbb{T}_{s2w} &= T_{s2w} \times R_{S} \times T^{-1}_{s2w} \\
    &= T^{-1}_{w2s} \times R_{S} \times T_{w2s} \\
    &= T^{-1}_{w2s} \times R^{left}_{X}(p) \times R^{left}_{X}(y) \times R^{left}_{Z}(r) \times T_{w2s} ,
\end{split}
\end{equation}
where $T_{w2s}$ and $y, p, r, R^{left}_{X}, R^{left}_{Y}, R^{left}_{Z}$ are defined in Formula \ref{formula:56} and \ref{eq:16} respectively.

\subsection{Error measurements for the pose conversion}
\begin{figure}[H]
\centering
\includegraphics[width=00.9\textwidth]{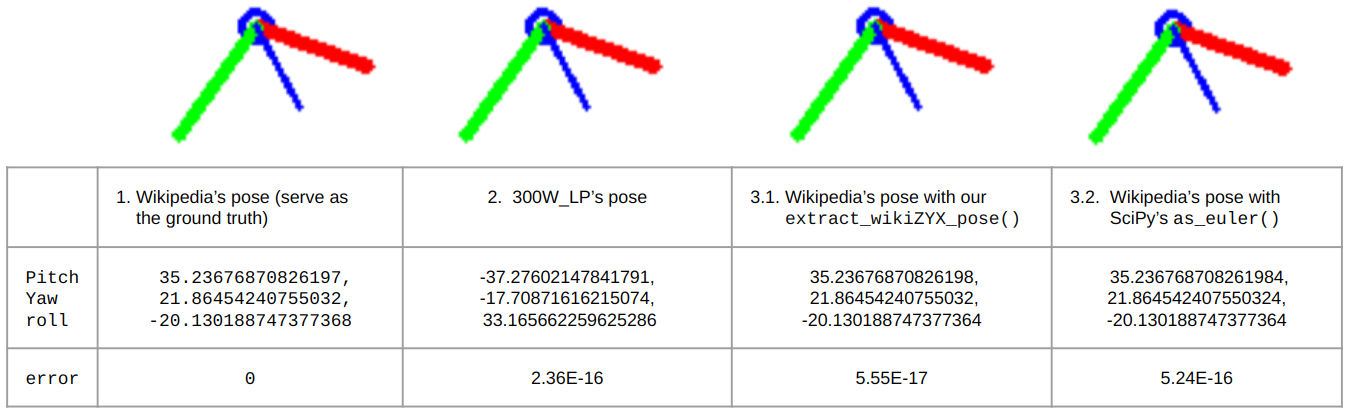}
\caption{There are 4 different poses in this figure. The first column shows the ground-truth pose (Pose 1) defined in SciPy/Wikipedia's rotation system. The second column shows the converted pose (Pose 2) in 300W{\_}LP's rotation system and the Frobenius norm between matrices extracted from Pose 1 and 2 in SciPy's coordinate system. The third column describes the converted pose (Pose 3.1), which is obtained by calling our extract{\_}wiki{\_}pose(), transformed from Pose 2, and the Frobenius norm between matrices extracted from Pose 1 and 3.1. The last column describes the converted pose (Pose 3.2), which is obtained by calling SciPy's as{\_}euler(), transformed from Pose 2, and the Frobenius norm between matrices extracted from Pose 1 and 3.2. We used our three-line drawing routines for Wikipedia and 300W{\_}LP's rotation systems to plot the rotations.}
\label{fig:pose_conversion}
\end{figure}

The conversion of Euler angles between different coordinate/rotation systems inevitably introduces numerical errors.
Error/loss measuring can be done directly for poses in the same coordinate system but can't be directly computed through poses of two different rotation systems.  We measure the pose conversion through a matrix norm between the two rotation matrices in the same coordinate system to avoid dealing with different axes-sequences and coordinate systems. So, to compare with the rotation matrix of the first pose, we need to reverse the process described in Sec. \ref{sec:pose_conversion}, transforming the converted pose in the second rotation system back to its equivalent rotation matrix in the first system.  The rotation matrices can be measured through the Frobenius or Geodesic norms. Both recognize more subtle differences between rotations. For simplicity, we use the Frobenius norm.  Fig. \ref{fig:pose_conversion} demonstrates our conversions between SciPy and 300W{\_}LP's poses.

%\section{Rotation system defined in 3DDFA{\_}v2's implementation \cite{DBLP:conf/cvpr/ZhuLLSL16}}
\section{Inferring 3DDFA{\_}v2's rotation system from its implementation \cite{DBLP:conf/cvpr/ZhuLLSL16}}
\label{sec:3ddfa}
Both 3DDFA \cite{3ddfa_cleardusk} and 3DDFA{\_}v2 \cite{DBLP:conf/eccv/GuoZYYLL20} are the improved versions of the original paper \cite{DBLP:conf/cvpr/ZhuLLSL16}, and evaluated 300W-LP based on their implementation.  Though sharing the same names,  the 3DDFA, an abbreviation of the paper \cite{3ddfa_cleardusk}, is not the algorithm 3DDFA mentioned in \cite{DBLP:conf/cvpr/ZhuLLSL16}. Their implementations are similar: first, apply Dlib or FaceBoxes \cite{zhang2018faceboxes} face detectors to extract the regions of heads, then infer the 3DMM parameters from the pre-trained models for each bounding region. To achieve fast inference, 3DDFA v2 clips the 3DMM \cite{1227983} model parameters, $\mathbf{\alpha_{id}} \in \mathbb{R}^{199}$ and $\mathbf{\alpha_{exp}} \in \mathbb{R}^{29}$, into the first 40 and 10 dimensions respectively, and reduces the parameters to ${\mathbf{p} = [T^{3 \times 4}, \mathbf{\alpha}^{50}]}$. In the following sub-section, we'll explain how $T^{3 \times 4}$ works.

% \subsection{Inferring 3DDFA{\_}v2's rotation system}
3DDFA{\_}v2 \cite{DBLP:conf/eccv/GuoZYYLL20} uses the 300W-LP dataset in its experiments.  So, to derive 3DDFA{\_}v2's Euler angles for the 300W-LP dataset, we carefully examined its implementation \cite{DBLP:conf/cvpr/ZhuLLSL16} and found out that, for each detected bounded box, the pre-trained TDDFA model will output $T^{3 \times 4}$. $T^{3 \times 4}$ then can be decomposed into a 3x3 rotation matrix, $R_{W'}$, a scaling factor $s$, and a translation offset $t3d$. We denote 3DDFA{\_}v2's rotation matrix by $R_{W'}$ in case this definition is different from 300W{\_}LP's $R_{W}$ notation. We notice that 3DDFA{\_}v2 applies the pose frustum drawing scheme discussed in Sec. \ref{300wlp_euler_drawing} (See Fig. \ref{fig:pose_box}) without really using its yaw, roll, pitch values. 

The function \textbf{matrix2angle}() in 3DDFA{\_}v2/utils/pose.y (shown in Listing \ref{listing:3}) is the only source code to provide the yaw-roll-pitch extraction for $R_{W'}$. 
% The Docstrings of \textbf{matrix2angle}() specifies the relation between axes and Euler angles, the X-, Y-, and Z-axes associated with yaw, pitch, and roll, respectively. 
The discussion about the Base Face Model (BFM) in Sec. \ref{sec:300wlp-fp} implies both 3DDFA and 3DDFA{\_}v2 and 300W{\_}LP share the same coordinate system. Don't be misled by the phrase, \textit{x(yaw), y(pitch), and z(roll)}, in \textbf{matrix2angle}()'s Docstrings because they don't mean the associated rotation really rotating along the X-, Y-, and Z-axes respectively. 

The non-gimbal-lock case (Lines 23-25) of \textbf{matrix2angle}() encourages the first column and first row of $R_{W'}$ to follow the pattern (\ref{formula:1}) where $y,p, r$ means yaw, pitch and roll respectively:
\begin{equation}
 \label{formula:1}
R_{W'}(y,p,r) = 
\begin{bmatrix}
cos(r)cos(y)  & - & -\\
sin(r)cos(y)  & - & -\\
sin(y)  &  cos(y)sin(p)   &  cos(y)cos(p)
\end{bmatrix}
\end{equation}
Because each Euler angle rotates along one of the X-, Y-, or Z-axes and different Euler angles rotate along different axes,  the three fundamental rotation matrices associated with yaw, roll, and pitch have the zero-one column and row distributions (See (\ref{eq:2})).  Once the zero-one pattern is fixed for one Euler angle, say yaw, the other Euler angles won’t have the same zero-one pattern.   All possible fundamental rotation matrices are shown in (\ref{eq:3}).  
\begin{equation}
 \label{eq:2}
\begin{bmatrix}
1  & 0 & 0\\
0  & - & -\\
0  &  -   &  -
\end{bmatrix},
\begin{bmatrix}
-  & 0 & -\\
0  & 1 & 0\\
-  &  0   &  -
\end{bmatrix},
\begin{bmatrix}
0  & 0 & 1\\
-  & - & 0\\
-  &  -   &  0
\end{bmatrix}
\end{equation}
\begin{equation}
 \label{eq:3}
\begin{bmatrix}
1  & 0 & 0 \\
0  & cos(\theta) & 	 \mp sin(\theta)\\
0  & \pm sin(\theta)   &  cos(\theta)
\end{bmatrix},
\begin{bmatrix}
cos(\theta)  & 0 & \pm sin(\theta)\\
0  & 1 & 0\\
\mp sin(\theta)  &  0   &  cos(\theta)
\end{bmatrix},
\begin{bmatrix}
cos(\theta) & 	 \mp sin(\theta) & 0\\
\pm sin(\theta)   &  cos(\theta)   &  0 \\
0  & 0 & 1\\
\end{bmatrix}
\end{equation}
We compute the yaw-roll-pitch rotation matrices from all possible fundamental rotation matrix multiplications to find the solutions for $R_{W'}(y,p,r)$ with the given pattern (\ref{formula:1}) used in 3DDFA{\_}v2. Among all $\pm$ y, $\pm$p, $\pm$r, X, Y, Z choices, there’s one unique solution found (Formula (\ref{eq:5})), and we use the notation $R^{right}_{Z}(r)$ to represent the roll rotation along the Z-axis in the right-handed manner in 300W{\_}LP's coordinate system.
\begin{equation}
\label{eq:4}
R^{right}_{X}(p) = 
\begin{bmatrix}
1 & 0 & 0 \\
0 & cos(p) & -sin(p) \\
0 & sin(p) &  cos(p)
\end{bmatrix}, \\ 
R^{left}_{Y}(y) = 
\begin{bmatrix}
cos(y) & 0 & -sin(y) \\
0  & 1 & 0\\
sin(y) & 0 & cos(y)
\end{bmatrix},
R^{right}_{Z}(r) = 
\begin{bmatrix}
cos(r)  & -sin(r) & 0\\
sin(r) & cos(r) &  0\\
0   &     0      &  1
\end{bmatrix},
\end{equation}
\begin{equation}
\label{eq:5}
\begin{split}
R^{'}_{W}(y, p, r) &= R^{right}_{Z}(r) \times R^{lefft}_{Y}(y) \times R^{right}_{X}(p) \\
&= \begin{bmatrix}
cos(r)cos(y)  &   -sin(p)sin(y)cos(r) - sin(r)cos(p)   &  sin(p)sin(r) - sin(y)cos(p)cos(r) \\
sin(r)cos(y)  &  - sin(p)sin(r)sin(y) + cos(p)cos(r))  &  -sin(p)cos(r) - sin(r)sin(y)cos(p)\\
sin(y)       &         sin(p)cos(y)           &         cos(p)cos(y) 
\end{bmatrix} 
\end{split}
\end{equation}
 So we show that Formula (\ref{eq:5}) is the unique solution to Eq. (\ref{formula:1}). Comparing 3DDFA and 3DDFA{\_}v2's $R^{'}_{W}(y, p, r) = R^{right}_{Z}(r) \times R^{lefft}_{Y}(y) \times R^{right}_{X}(p)$ with 300W{\_}LP's $R_{W}(y, p, r) = R^{left}_{X}(p) \times R^{left}_{Y}(y) \times R^{left}_{Z}(r)$,   we proved that 3DDFA and 3DDFA{\_}v2 uses different axis-sequence and handedness from 300W{\_}LP's. 

 To align with 300W{\_}LP's rotation system and adjust \textbf{matrix2angle}'s errors, we suggest applying the rotation formula $R_{W}$ in Formula \ref{eq:16}, the pose extraction from Listing \ref{listing:1}, and one of the 300W{\_}LP's drawing routines mentioned in Sec \ref{our_300wlp_drawing_section}.
\begin{figure}[H]
\centering
\includegraphics[width=0.5 \textwidth]{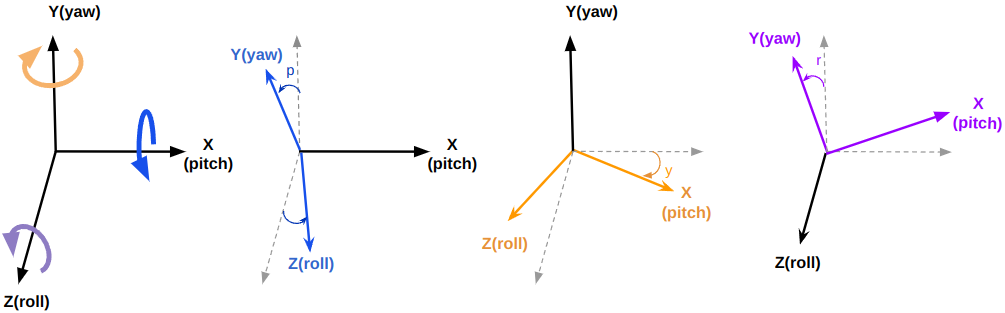}
\caption{\textbf{matrix2angle} defines mix-handed elemental rotations}
\label{fig:3ddfa_euler_angles}
\end{figure}

\section{Head pose estimation: 6D-RepNet \cite{Hempel_2022} and 6D-RepNet360 \cite{hempel2023robust}}
\label{sec:6drepnet}
6D-RepNet uses a network RepVGG \cite{ding2021repvgg}, allowing skip-connection-like structure during training time while the skip connections are merged into conv-net operation during inference.  Deep learning generates features by itself, so there is no need for key-point detection.  Both 6D-RepNet and 6D-RepNet360 introduce a 6D representation as the neural network's output. Subsequently, the Gram-Schmidt process is employed to construct a rotation matrix from the 6D representation, with learning facilitated through the geodesic loss of rotation matrices. However, both 6D-RepNets' papers \cite{Hempel_2022} and \cite{hempel2023robust} lack explicit statements about its rotation system used to train/test with 300W-LP alone or combining with CMU Panoptic. Since there's no coordinate transformation involved, we can only assume 6D-RepNet inherits 300W{\_}LP's coordinate system as it is training directly with 300W{\_}LP's ground truth labels.

Both 6D-RepNets interpret the yaw, roll, and pitch from the predicted rotation matrix through the function shown in Listing \ref{listing:4}. Then, plot the Euler angles with the three-line and pose-box drawing routines, as shown in Listing \ref{listing:5} and \ref{listing:6}.  We must consider them all to unveil 6D-RepNet's rotation system and Euler angle definitions.
% \begin{figure}
%   \begin{minipage}[c]{0.3\linewidth}
%   \includegraphics[width=\textwidth,height=3cm]{figures/6drep.jpg}
%   \caption{6DRepNet's Demo}
%   \label{fig:6d1}
%   \end{minipage}
%   \hfill
%   \begin{minipage}[c]{0.65\linewidth}
  
%   \includegraphics[width=\textwidth,height=3cm]{figures/6d.png}
%   \caption{6DRepNet's proposed method}
%   \label{fig:sub2}
%   \end{minipage}%
% \label{fig:6d}
% \end{figure}

\subsection{6D-RepNet's rotation system vs 300W{\_}LP's: are they the same?}
Let's start with its drawing routines,  \textbf{draw{\_}axis}() in Listing \ref{listing:5} and \textbf{plot{\_}pose{\_}cube}() in Listing \ref{listing:6}. Because \textbf{plot{\_}pose{\_}cube}() follows \textbf{draw{\_}axis}()'s logic, we only need to consider \textbf{draw{\_}axis}(). Sec. \ref{sec:300wlp_3line_drawing} has shown that \textbf{draw{\_}axis}()'s three lines (Formula \ref{formula:45}) correspond to 
\begin{equation}
\label{formula:62}
\begin{split}
R^{draw}_{img}(p, y, r) &= \begin{bmatrix}
1 & 0 & 0 \\ 0 & 1 & 0    
\end{bmatrix} \times R^{draw}_{W}(p, y, r) \\
&= \begin{bmatrix}
    cos(y)cos(r) & -cos(y)sin(r) & -sin(y) \\
    cos(p)sin(r)-sin(p)sin(y)cos(r) & cos(p)cos(r)+sin(p)sin(y)sin(r) & -sin(p)cos(y)
\end{bmatrix}.    
\end{split}
\end{equation}

\begin{lemma} \label{thm:drawaxis}
 $R^{draw}_{W}(p, y, r)$ is the only rotation matrix satisfying 
\begin{equation}
\label{formula:63}
R^{draw}_{img} = \begin{bmatrix}
1 & 0 & 0 \\ 0 & 1 & 0    
\end{bmatrix} \times R, \quad where \,\; R \,\; is \,\; a \,\;  3 \times 3 \,\; rotation \,\; matrix.  
\end{equation}
\end{lemma}
\begin{proof}
Since $R^{draw}_{W}(p, y, r)$ is the rotation matrix (check Sec. \ref{our_300wlp_drawing_section}), and let $\vec{a}$ and $\vec{b}$ represent the first and second rows of $R^{draw}_{W}(p, y, r)$, $\vec{a}$ and $\vec{b}$ must be orthonormal.  The fact of $R$'s determinant $=1$ leads to the third row of $R$ equal to the cross product of $\vec{a}$ and $\vec{b}$, i.e., $\vec{a} \times \vec{b}$.  So, we proved that only one unique rotation matrix satisfies Eq. \ref{formula:63}. Then $R = R^{draw}_{W}(p, y, r)$. 
\end{proof}
Combining Lemma \ref{thm:drawaxis} with Formula \ref{formula:30}, we can derive 6D-RepNet's rotation, denoted by $R^{*}_{6d}$, equals $R_{W}$ as shown below \footnote[1]{We use the notations $R^{handedness}_{axis}$ to represent the specified left/right-handed rotations along the specified axis in 300W{\_}LP's coordinate system. For example, $R^{left}_{X}$ represents the left-handed rotation along the X-axis.}. 
\begin{equation}
\label{formula:64}
\begin{split}
    R^{*}_{6d} &= T^{-1}_{W} \times R^{draw}_{W}(p, y, r) \times T_{W} = R_{W}\\        
    \implies R^{*}_{6d} &= R_{W} = R^{left}_{X}(p) \times R^{left}_{Y}(y) \times R^{left}_{Z}(r).
\end{split}
\end{equation}
Thus \textbf{draw{\_}axis}() completely follows 300W{\_}LP's rotation system.

Next, let's dive into 6D-RepNet's rotation matrix formula, Formula \ref{formula:60} and \ref{formula:61} implemented in \textbf{get{\_}R}() (check Listing \ref{listing:4} for details). 
\begin{equation}
\label{formula:60}
R^{right}_{X}(p) = 
\begin{bmatrix}
1 & 0 & 0 \\
0 & cos(p) & -sin(p) \\
0 & sin(p) &  cos(p)
\end{bmatrix}, \\ 
R^{right}_{Y}(y) = 
\begin{bmatrix}
cos(y) & 0 & sin(y) \\
0  & 1 & 0\\
-sin(y) & 0 & cos(y)
\end{bmatrix},
R^{right}_{Z}(r) = 
\begin{bmatrix}
cos(r)  & -sin(r) & 0\\
sin(r) & cos(r) &  0\\
0   &     0      &  1
\end{bmatrix},
\end{equation}
\begin{equation}
\label{formula:61}
\begin{split}
R^{**}_{6d}(y, p, r) &= R^{right}_{Z}(r) \times R^{right}_{Y}(y) \times R^{right}_{X}(p) \\
&= \begin{bmatrix}
cos(r)cos(y)  &   sin(p)sin(y)cos(r) - sin(r)cos(p)   &  sin(p)sin(r) + sin(y)cos(p)cos(r) \\
sin(r)cos(y)  &  sin(p)sin(r)sin(y) + cos(p)cos(r)  &  -sin(p)cos(r) + sin(r)sin(y)cos(p)\\
-sin(y)       &         sin(p)cos(y)           &         cos(p)cos(y) 
\end{bmatrix} 
\end{split}
\end{equation}
It shows another rotation matrix formula, denoted as $R^{**}_{6d}$, $R^{**}_{6d} = R^{right}_{Z}(r) \times R^{right}_{Y}(y) \times R^{right}_{X}(p)$ , which contradicts $R^{*}_{6d}$ of Formula \ref{formula:64}.  
Therefore, 6D-RepNet applies 2 rotation systems defined in 300W{\_}LP's coordinate system, $R^{**}_{6d}$ for head pose dataset retrieval/training/inference, $R^{*}_{6d}$ for the drawing purpose. Luckily, we notice that $R_{W}$ and $R^{**}_{6d}(y, p, r)$ are inverse of each other, i.e.,
\begin{equation}
\label{formula:65}
  \begin{split}
     R^{**}_{6d}(y, p, r) &= R^{right}_{Z}(r) \times R^{right}_{Y}(y) \times R^{right}_{X}(p) \\\
                          &= R^{left}_{Z}(-r) \times R^{left}_{Y}(-y) \times R^{left}_{X}(-p) \\
                          &= (R^{left}_{X}(p) \times R^{left}_{Y}(y) \times R^{left}_{Z}(r))^{-1} \\
                          &= R_{W}(y, p, r)^{-1} \\
                          &= R_{W}(y, p, r)^{transpose}.
  \end{split}
\end{equation}
The above derivation implies two observations. Firstly, 6D-RepNet's neural network learning with $R^{**}_{6d}(y, p, r)$-rotations still works because the Geodesic distance (used in 6D-RepNet's training) between two $R^{**}_{6d}$-rotations remains the same as the Geodesic distance between their inverse rotations, i.e., $R_{W}$-rotations. Secondly, \textbf{draw{\_}axis}() works for 6D-RepNet's Euler-angle predictions because $R^{**}_{6d}$'s predicted pitch, yaw, roll also remain the same as $R_{W}$'s according to Equality \ref{formula:65}. 

Finally, we notice that 6D-RepNet's Euler angle extraction, shown in Listing \ref{listing:12}, only outputs one set of pitch, yaw, and roll.  As discussed at the end of Sec. \ref{subsec:extract_300w_euler}, if we can show yaw $\in [-90\degree, 90\degree]$, then 6D-RepNet and 300W{\_}LP share the same pose selection.  Because Listing \ref{listing:12}'s Line 7's $sy = |cos(yaw)|$ forces Line 12, $y = torch.atan2(-R[:, 2, 0], sy)$, to output yaw $\in [-90\degree, 90\degree]$, 6D-RepNet and 300W{\_}LP indeed share the same pose selection (the first one as in Listing \ref{listing:1}).

\subsection{Conclusion}
6D-RepNet adapts 300W{\_}LP's inverse rotation system. So, while its predicted rotation matrices don't align with 300W{\_}LP's rotation system, its extracted Euler angles coincide with 300W{\_}LP's Euler angles. Hence, we can plot the predicted 300W{\_}LP's rotations correctly by plugging the predicted Euler angles into \textbf{draw{\_}axis}().

\section{\href{https://github.com/Ascend-Research/HeadPoseEstimation-WHENet/blob/master/prepare_images.py}{WHENet's head pose generation}}
To train its HPE model, WHENet \cite{zhou2020whenet} follows the convention of training with 300W{\_}LP and testing with AFLW2000 \cite{DBLP:conf/cvpr/ZhuLLSL16} and BIWI \cite{DBLP:journals/ijcv/FanelliDGFG13}. Although 300W{\_}LP contains 150K images, it lacks images with yaws close to 0 or outside of the angle range $[-90\degree, 90\degree]$. On the other hand, the CMU Panoptic Dataset \cite{Joo_2017_TPAMI} captures multiple (occlusive) subjects from an abundance of calibrated cameras covering a full hemisphere and provides facial landmarks in 3D. WHENet, as far as we can tell, was the first to propose an open-sourced code for their pose extraction method, implemented in prepare{\_}images.py (Listing \ref{listing:11}), for the CMU Panoptic Dataset \cite{Joo_2017_TPAMI}. They estimated head pose from the provided facial keypoints and calculated rotations to go from the standard-pose face model to it using Horn's closed-form formula \cite{article3}. 
WHENet then combines the frontal-poses and anterior-view pose labels from 300W{\_}LP and CMU Panoptic Datasets to construct a more balanced head pose dataset with the desired full-pose-range coverage.

\href{https://github.com/Ascend-Research/HeadPoseEstimation-WHENet/blob/master/utils.py}{WHENet} \cite{zhou2020whenet} also uses \textbf{draw{\_}axis}() (Listing \ref{listing:5}) for three-line drawing. As discussed in Sec. \ref{sec:6drepnet}, \textbf{draw{\_}axis} is tailored for the Euler angles from 300W{\_}LP's rotation system. So, the rest of this section will focus on whether WHENet's head pose generation for CMU Panoptic Dataset follows 300W{\_}LP's rotation system or not.

\subsection{The head pose generation for CMU Panoptic Dataset}
\label{whenet_pose_gen}
The head pose generation is implemented in the function \href{https://github.com/Ascend-Research/HeadPoseEstimation-WHENet/blob/master/prepare_images.py}{\textbf{save{\_}img{\_}head}}(), i.e., Listing \ref{listing:11}, and needed to be examined.
\begin{figure}[H]
\centering
\includegraphics[width=0.3\textwidth]{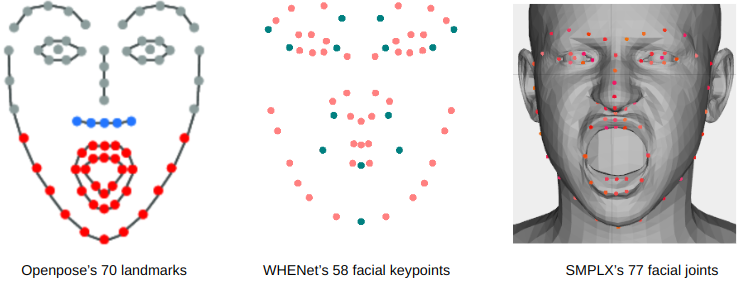}
\caption{From left to right: \href{https://www.researchgate.net/figure/Mask-refinement-via-OpenPose-landmarks-OpenPose-body-landmarks-left-and-facial_fig3_363074737}{Openpose's 70 lanmarks}, WHENet's 58-keypoint BFM, and SMPLX's \cite{pavlakos2019expressive} 77 face joints.}
\label{fig:whenet_bfm_vs_openpose}
\end{figure} 
\subsubsection{Coordinate systems: WHENet v.s. 300W{\_}LP}
\label{whenet_coord_system}
CMU Panoptic Dataset uses OpenPose's \cite{8765346} 70 facial landmarks \cite{simon2017hand} to define each individual's unique 3D face.  Due to no corresponding face model for the \textbf{standard head pose}(head pose without any rotation, i.e., yaw=pitch=roll=0),  WHNENet uses the 58-keypoint face model $X_{ref}$  defined by Listing \ref{listing:8} to represent the standard-pose face model(the middle in Fig. \ref{fig:whenet_bfm_vs_openpose}).  Due to the landmark difference, WHENet picks 6-14 keypoints (green dots in Fig. \ref{fig:whenet_bfm_vs_openpose}) and applies Horn's closed-form solution \cite{article3} to estimate the rotation $R_{Horn}$ (i.e., $temp$ in the code) from the keypoints to the corresponding CMU Panoptic Dataset's ground-truths. For simplicity, we assume $R_{Horn}$ and $X_{ref}$ are $3 \times 3$. We find out that while $X_{ref}$ follows 300W{\_}LP's coordinate system (left in Fig. \ref{fig:300w-lp-system}), OpenPose's standard-pose landmarks turn $180\degree$  along the X-axis. From Lines 35 - 45 in Listing \ref{listing:11}, we found that WheNet tries to eliminate such an impact by applying the coordinate transformation associated with $E_{ref}$. Its rotation matrix output is as follows: 
\begin{equation}
    \label{eq:50}
    \begin{split}
    &E_{ref} =\begin{bmatrix}
            1 & 0 & 0 \\
            0 & -1 & 0 \\
            0 & 0 & -1
          \end{bmatrix}, 
    CR = C_{extr} \times R_{Horn}, \\
    &R_{whenet} \coloneqq E_{ref} \times CR \times E_{ref}^{-1}.
    \end{split}
\end{equation}
Note that $C_{extr}$ is the camera extrinsic (i.e., cam['R'] in Listing \ref{listing:11}) provided by the Panoptic dataset.  But from our understanding and experiments, $CR$ in Formula \ref{eq:50} is already a rotation defined in the 300W{\_}LP's coordinate system. Refer to Sec. \ref{sec:whenet_300wlp} for more details.

\subsubsection{Rotations: WHENet v.s. 300W{\_}LP} \label{sec:whenet_300wlp}
To effectively eliminate the difference between Whenet's standard head pose ($X_{ref}$) and OpenPose's, we improved the rotation matrix formula from Formula \ref{eq:50} for the Panoptic dataset to the following: 
\begin{equation}
    \label{formula:66}
    R_{panoptic} \coloneqq E_{ref} \times C_{extr} \times R_{Horn}
\end{equation}
\begin{figure}[H]
\centering
\includegraphics[width=0.9\textwidth]{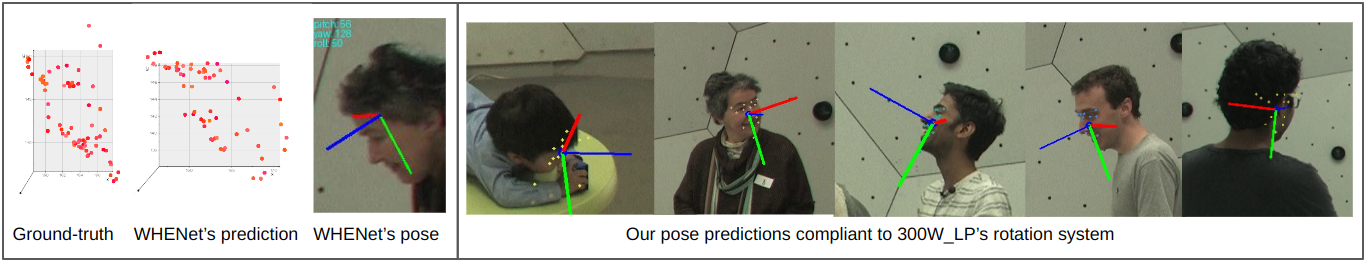}
\caption{Left panel shows WHENet's predicted facial landmarks (middle) and head rotation (right) based on ground-truth landmarks (left). The red line should stretch to the right, not the left, so the predicted pose is not desired. On the contrary, Right panel shows the proposed Formula \ref{formula:66} greatly improves the the head pose prediction in 300W{\_}LP's system!  }
\label{fig:cmu}
\end{figure} 
\begin{figure}[H]
\centering
\includegraphics[width=0.7\textwidth]{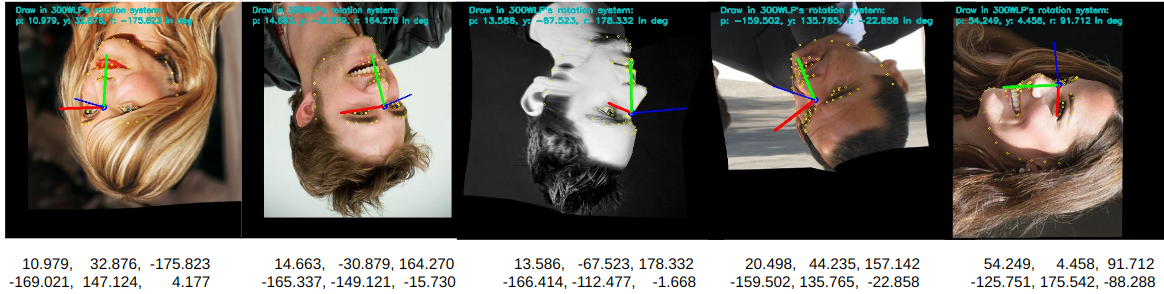}
\caption{WheNet's select{\_}euler() in Listing \ref{listing:0} follows 300W{\_}LP's Euler-angle extraction method with one constraint: keeping pitch and roll values $\in [-90\degree, 90\degree]$! It leads to yaw beyond $\pm90\degree$ sometimes. If pitch or roll isn't in this range, -999 will be assigned. The last 2 rows show the complete pitch-yaw-roll solutions for the above images. Note that yaws are within $[-90\degree, 90\degree]$ and beyond $\pm90\degree$ for the first and second rows respectively. While 6D-RepNet's Euler-angle extraction outputs the first row as the head pose solution, WHENet's outputs $(-999, -999, -999)$.}
\label{fig:euler_angle_diff}
\end{figure} 

\subsection{The head pose generation for the AGORA dataset \cite{patel2021agora}} 
\href{https://agora.is.tue.mpg.de/}{AGORA} \cite{patel2021agora} is a synthetic dataset which places various SMPL-X \cite{pavlakos2019expressive} body models on realistic 3D environments. It has around 14K training and 3K test images, rendering between 5 and 15 people per image. In total, AGORA consists of 173K person crops. 
DirectMHP's (DMHP) \cite{zhou2023directmhp}'s head pose generation for the AGORA dataset, \href{https://github.com/hnuzhy/DirectMHP/blob/main/exps/AGORA/data_process_hpe.py}{auto{\_}labels{\_}generating()} in Listing \ref{listing:7}, follows WHENet's save{\_}img{\_}head()(Listing \ref{listing:11}) with the AGORA's 13 face joints selection and the face model tilted $X_{ref}$ $10\degree$ along the X-axis. So, we conclude that DMHP completely adapts WHENet's rotation system and Euler angle extraction.

%\section{Head pose dataset augmentation in 300W{\_}LP's rotation system}
\section{2D image data augmentation for head pose dataset}
This section will focus on the standard 2D geometric image augmentation commonly employed in deep-learning-based image classification, such as rotations/flipping and their effects on rotation matrices. Data augmentation is a crucial tool in the training pipeline for computer vision models, enabling them to achieve higher performance, improved generalization, and greater robustness without increasing the number of training images. However, in HPE, geometric image augmentations have been avoided due to a lack of clear definition of rotation systems and mathematical derivation of new rotation matrices under 2D geometric transformation. Clearly, geometrical transformation also enhances angle coverage for free.

It is easy to observe that pixel-level image augmentations do not alter the rotation matrices. Readers can easily compute the Euler angles with knowledge from Sec. \ref{subsec:extract_300w_euler} when the correct rotation matrices are available. 

\begin{defn} \label{aug_type_def} The 2D images are assumed to be on the $XY$-plane of the 300W{\_}LP's coordinate system as in Figure \ref{fig:300w-lp-system}. 2D images' \textbf{horizontal \& vertical flipping} are the flipping across the image's vertical Y-axis and horizontal X-axis, respectively. On the other hand, to align with OpenCV's built-in affine transformation, we define a \textbf{image's rotation with the angle $\bm{\phi}$} to be the rotation rotating the image an angle, $\phi$, counter-clockwise. 
\end{defn} 

\begin{defn} \label{L_def}
Define $\bm{L_{\theta}}$ to be the line passing through the origin and the angle from the horizontal axis and itself (on the vertical plane, denoted by $\bm{P_{v}}$, intersecting with the horizontal and vertical axes) is $\theta$. For example, the line $L_{\theta}$ of 300W{\_}LP's coordinate system represents one constructed from rotating the X-axis ($y = 0$) by the angle $\theta$ counter-clockwise on the XY-plane.  Because of 300W{\_}LP's left-handed nature for the XY-plane's rotations, such rotations are clockwise. So $\theta > 0$ is equivalent to the roll rotation $r = - \theta$ in 300W{\_}LP's rotation system. However, Wikipedia's rotation system's right-handed nature guarantees its roll rotation $r = \theta$ will remain the same sign as $L_{\theta}$'s $\theta$.
\end{defn}

\begin{defn} \label{gen_flip_def}
Next, we define the \textbf{2D flipping about $\bm{L_{\theta}}$} to be the flipping across $L_{\theta}$ on the vertical plane $\bm{P_{v}}$ defined in Definition \ref{L_def}. 
\end{defn}

We will derive formulas under 300W{\_}LP's coordinate system and conventions. Similar derivations can be carried out easily for other coordinate systems.
\begin{theorem} \label{thm:2d_rotation}
Fix a 3D rotation system. Suppose an image contains a human head, and the head's intrinsic rotation $R$ in the rotation system is given. Prove that the 3D rotation, denoted by $R_{rotate{\_}img}(\phi)$, associated with rotating this image by an angle $\phi$ is the same to apply the extrinsic rotation, denoted as $R_{extr}(\phi)$, of rotating the axis (perpendicular to the image) by the angle $\phi$ on $R$. Hence, we have 
\begin{equation}
  \label{formula:58}
    R_{rotate{\_}img}(\phi) = R_{extr}(\phi)  \times R      
\end{equation}
\end{theorem}

\begin{figure}[H]
\centering
\includegraphics[width=0.85 \textwidth]{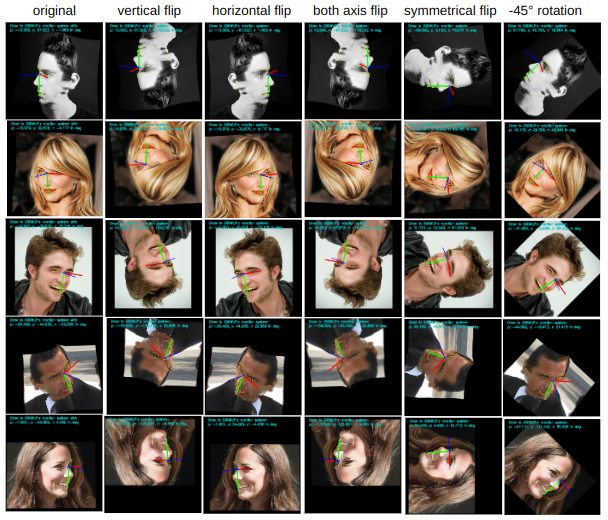}
\caption{Our pose \& rotation augmentations for the 300W{\_}LP dataset}
\label{fig:pose_augmentation}
\end{figure}

\begin{proof}
Without loss of generality, let's fix to 300W{\_}LP's rotation system and $R \ne I_{3}$.  Due to 300W{\_}LP's clockwise-roll nature, rotating the image by $\phi$ counter-clockwise results in the roll's elemental rotation $R^{left}_{Z}(-\phi)$ defined in Formula \ref{eq:16}. Moreover, this rotation is extrinsic because the image's rotation can only happen on 300W{\_}LP's extrinsic $xy$-plane. Then we apply Lemma \ref{thm:diff_rots} and can derive that $R_{rot{\_}img}(\phi)$ satisfies Formula \ref{eq:39}.
\begin{equation}
  \label{eq:39}
  \begin{split}
    &R_{extr}(\phi) = R^{left}_{Z}(-\phi), \\
    &R_{rot{\_}img}(\phi) = R_{extr}(\phi) \times R = R^{left}_{Z}(-\phi) \times R \\ \implies 
    &R_{rot{\_}img}(\phi)
    =  \begin{bmatrix}
        cos(\phi) & -sin(\phi) & 0\\
        sin(\phi) & cos(\phi) & 0\\
        0 & 0 & 1
       \end{bmatrix} \times R 
  \end{split}
\end{equation}
\end{proof}
\begin{theorem} \label{thm:flip}
Let's fix 300W{\_}LP's rotation system and restrict $0\degree \leq \theta \leq 90\degree$. Suppose an image contains a human head, and the head's intrinsic rotation $R$ in the fixed rotation system is given. Prove that the 3D rotation $R_{flip}(\theta)$, corresponding to the 2D flipping across $L_{\theta}$ and the given head's rotation $R$, can be expressed as follows:
\begin{equation}
  \label{formula:52}
    R_{flip}(\theta) = \begin{bmatrix}
        cos(2\theta) & sin(2\theta) & 0\\
        sin(2\theta) & -cos(2\theta) & 0\\
        0 & 0 & 1
       \end{bmatrix}  \times R \times \begin{bmatrix}
        -1 & 0 & 0\\
        0 & 1 & 0\\
        0 & 0 & 1
       \end{bmatrix}     
\end{equation}
\end{theorem}
\begin{proof}
Without loss of generality, let $R \ne I_{3}$. It's trivial that the 2D flipping across $L_{\theta}$ will flip from $\bm{e_{x}} \coloneqq (1, 0, 0)$ to $(cos(2\theta), sin(2\theta), 0)$. But the 2D flipping across $L_{\theta}$ flips $\bm{e_{y}} \coloneqq (0, 1, 0)$ to $(sin(2\theta), -cos(2\theta), 0)$ requires some work. 
So the matrix of the extrinsic 3D flipping is $\begin{bmatrix}
        cos(2\theta) & sin(2\theta) & 0\\
        sin(2\theta) & -cos(2\theta) & 0\\
        0 & 0 & 1
       \end{bmatrix}$. 
The flipping also causes the intrinsic X-axis to change to the opposite direction, which results in the intrinsic flipping across the X-axis, i.e., 
$\begin{bmatrix}
-1 & 0 & 0 \\
0 & 1 & 0 \\
0 & 0 & 1
\end{bmatrix}$.  Therefore, the 2D flipping across $L_{\theta}$ relates to the 3D rotation, which applies an extrinsic $L_{\theta}$-flipping and an intrinsic $X$-flipping. By Lemma \ref{thm:diff_rots}, we have $R_{flip}(\theta) =  (extrinsic{\_}L_{\theta}{\_}flipping) \times R  \times (intrinsic{\_}X{\_}flipping)$, and thus proved Formula \ref{formula:52}.
\end{proof}
%%%%%%%%%%%%%%%%%%%%%%%%%%%%%%%%%%%%start corollary
\begin{corollary}\label{cor:flipcor} Here we list a few special cases of the above theorem under 300W{\_}LP's rotation system:
    \begin{enumerate}
    \item Horizontal flipping is $R_{flip}(\pi/2) =  \begin{bsmallmatrix}
                        -1 & 0 & 0\\
                        0 & 1 & 0\\
                        0 & 0 & 1
                       \end{bsmallmatrix}  \times R \times \begin{bsmallmatrix}
                        -1 & 0 & 0\\
                        0 & 1 & 0\\
                        0 & 0 & 1
                       \end{bsmallmatrix}$.
    \item Vertical flipping is $R_{flip}(0) =  \begin{bsmallmatrix}
            1 & 0 & 0\\
            0 & -1 & 0\\
            0 & 0 & 1
           \end{bsmallmatrix}  \times R \times \begin{bsmallmatrix}
            -1 & 0 & 0\\
            0 & 1 & 0\\
            0 & 0 & 1
           \end{bsmallmatrix}$. 
    \item Flipping about both axes is
    $R_{bf} = \begin{bsmallmatrix}
            -1 & 0 & 0\\
            0 & 1 & 0\\
            0 & 0 & 1
           \end{bsmallmatrix}  \times \begin{bsmallmatrix}
            1 & 0 & 0\\
            0 & -1 & 0\\
            0 & 0 & 1
           \end{bsmallmatrix}  \times R \times \begin{bsmallmatrix}
            -1 & 0 & 0\\
            0 & 1 & 0\\
            0 & 0 & 1
           \end{bsmallmatrix} \times \begin{bsmallmatrix}
            -1 & 0 & 0\\
            0 & 1 & 0\\
            0 & 0 & 1
           \end{bsmallmatrix}= \begin{bsmallmatrix}
            -1 & 0 & 0\\
            0 & -1 & 0\\
            0 & 0 & 1
           \end{bsmallmatrix}  \times R.$  
    \item The symmetrical flipping about $\bm{L_{\pi /4}}$ is  
    $R_{flip}(\pi /4) = \begin{bsmallmatrix}
            0 & 1 & 0\\
            1 & 0 & 0\\
            0 & 0 & 1
           \end{bsmallmatrix}  \times R \times \begin{bsmallmatrix}
            -1 & 0 & 0\\
            0 & 1 & 0\\
            0 & 0 & 1
           \end{bsmallmatrix} .$
    \item The rotation $\pi/4$ counter-clockwise corresponds to
            $R_{rot}(\pi/4) = \begin{bsmallmatrix}
            cos(\pi/4) & -sin(\pi/4) & 0\\
            sin(\pi/4) & cos(\pi/4) & 0\\
            0 & 0 & 1
           \end{bsmallmatrix}  \times R$.     
\end{enumerate}  
\end{corollary} 
%%%%%%%%%%%%%%%%%%%%%%%%%%%%%%%%%%%%end corollary
Fig. \ref{fig:pose_augmentation} demonstrates Corollary \ref{cor:flipcor} applied to various head pose images.

%Bibliography
\bibliographystyle{unsrt}  
\bibliography{references}  

\begin{appendices}
\begin{figure}[H]
\centering
\includegraphics[width=0.4 \textwidth]{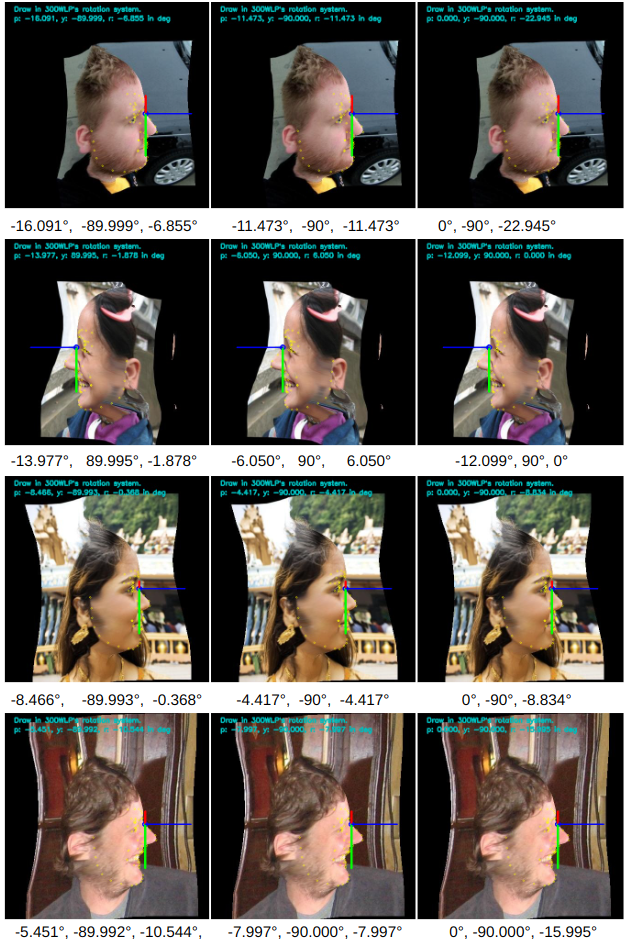}
\caption{300W{\_}LP's Gimbal-lock images: Euler angles follows pitch, yaw, and roll order. }
\label{fig:gimbol_lock2}
\end{figure} 

\begin{listing}[!ht]
\begin{minted}
[
frame=single,
framesep=2mm,
baselinestretch=1.2,
%bgcolor=LightGray,
%fontsize=\footnotesize,
linenos
]
{python}
y, p, r = rot_mtx.as_euler('ZYX', degrees=True)
\end{minted}
\caption{SciPy's Euler angle extraction}
\label{listing:9}
\end{listing}

\begin{listing}[!ht]
\begin{minted}
[
frame=single,
framesep=2mm,
baselinestretch=1.2,
%bgcolor=LightGray,
fontsize=\footnotesize,
linenos
]
{python}
        import numpy as np
        from np import arcsin, pi, arctan2, cos, sin, ndarray

        def extract_wikiZYX_pose(R_wiki: ndarray((3,3)):
            if np.abs(R_wiki[2, 0]) != 1.:
                p1 = -arcsin(R_wiki[2, 0])  
                p2 = pi - p1 if p1 >= 0 else -pi - p1  
                cp1 = cos(p1)
                r1 = arctan2(R_wiki[2, 1] / cp1, R_wiki[2, 2] / cp1)
                r2 = (r1 - pi) if r1 >= 0. else (r1 + pi)
                y1 = arctan2(R_wiki[1, 0] / cp1, R_wiki[0, 0] / cp1)
                y2 = (y1 - pi) if y1 >= 0. else (y1 + pi)
    
                return (p1, y1, r1), (p2, y2, r2)
            else:
                if R_wiki[2, 0] == -1:
                    p = pi / 2
                    r = arctan2(R_wiki[0, 1], R_wiki[0, 2]) / 2
                    y = -r
                else:
                    p = -pi / 2
                    y = r = arctan2(-R_wiki[0, 1], -R_wiki[0, 2]) / 2
                return (p, y, r), None
\end{minted}
\caption{Complete yaw, roll, pitch formula for Wikipedia's right-handed intrinsic ZYX-sequence rotations}
\label{listing:10}
\end{listing}

\begin{listing}
\begin{minted}
[
frame=single,
framesep=2mm,
baselinestretch=1.2,
%bgcolor=LightGray,
fontsize=\footnotesize,
linenos
]
{python}
        import numpy as np
        from np import arcsin, pi, arctan2, cos, sin, ndarray

        def extract_euler_angles_for_300W(R: ndarray((3,3)):
            half_pi = pi / 2
            y1 = arcsin(-R[0,2])
            if y1 == half_pi:
                p = arctan2(R[1,0], R[1,1]) / 2 
                r = -p
                return (p, y1, r)
            elif y1 == -half_pi:
                p = r = arctan2(-R[1,0], R[1,1]) / 2
                return (p, y1, r)
            else:  # non-Gimbal-lock cases
                cy1 = cos(y1)
                y2 = (pi - y1) if y1 >= 0 else (-pi - y1)
                p1 = arctan2(R[1,2] / cy1, R[2,2] / cy1 )
                p2 = (p1 - pi) if p1 >= 0 else (p1 + pi)
                r1 = arctan2(R[0,1] / cy1, R[0,0] / cy1)
                r2 = (r1 - pi) if r1 >= 0 else (r1 + pi)
                return (p1, y1, r1), (p2, y2, r2)
\end{minted}
\caption{Complete yaw, roll, pitch formula for the 300W-LP dataset}
\label{listing:1}
\end{listing}

% \begin{listing}[!ht]
% \begin{minted}
% [
% frame=single,
% framesep=2mm,
% baselinestretch=1.2,
% %bgcolor=LightGray,
% fontsize=\footnotesize,
% linenos
% ]
% {python}
%         import numpy as np
%         from np import arcsin, pi, arctan2, cos, sin, ndarray

%         def extract_euler_angles_for_3ddfa(R: ndarray((3,3)):
%             half_pi = pi / 2
%             y1 = arcsin(R[0,2])
%             if y1 == half_pi:
%                 p = r = arctan2(-R[0,1], -R[0,2]) / 2 
%                 return (p, y1, r)
%             elif y1 == -half_pi:
%                 p = arctan2(R[0,1], R[0,2]) / 2
%                 r = -p
%                 return (p, y1, r)
%             else:  # non-Gimbal-lock cases
%                 cy1 = cos(y1)
%                 y2 = (pi - y1) if y1 >= 0 else (-pi - y1)
%                 p1 = arctan2(R[2,1] / cy1, R[2,2] / cy1 )
%                 p2 = (p1 - pi) if p1 >= 0 else (p1 + pi)
%                 r1 = arctan2(R[1,0] / cy1, R[0,0] / cy1)
%                 r2 = (r1 - pi) if r1 >= 0 else (r1 + pi)
%                 return (p1, y1, r1), (p2, y2, r2)
% \end{minted}
% \caption{Complete yaw, roll, pitch formula for 3DDFA and 3DDFA{\_}v2's}
% \label{listing:2}
% \end{listing}

\begin{listing}[!ht]
\begin{minted}
[
frame=single,
framesep=2mm,
baselinestretch=1.2,
%bgcolor=LightGray,
% fontsize=\footnotesize,
linenos
]
{python}
def matrix2angle(R):
""" compute three Euler angles from a Rotation Matrix. Ref:             
    http://www.gregslabaugh.net/publications/euler.pdf
    refined by: https://stackoverflow.com/questions/43364900/
                rotation-matrix-to-euler-angles-with-opencv
    todo: check and debug
    Args:
        R: (3,3). rotation matrix
    Returns:
        x: yaw
        y: pitch
        z: roll
"""
if R[2, 0] > 0.998:
    z = 0
    x = np.pi / 2
    y = z + atan2(-R[0, 1], -R[0, 2])
elif R[2, 0] < -0.998:
    z = 0
    x = -np.pi / 2
    y = -z + atan2(R[0, 1], R[0, 2])
else:
    x = asin(R[2, 0])
    y = atan2(R[2, 1] / cos(x), R[2, 2] / cos(x))
    z = atan2(R[1, 0] / cos(x), R[0, 0] / cos(x))

return x, y, z
\end{minted}
\caption{3DDFA and 3DDFA{\_}v2's matrix2angle() function }
\label{listing:3}
\end{listing} 

\begin{listing}
\begin{minted}
[
frame=single,
framesep=2mm,
baselinestretch=1.2,
%bgcolor=LightGray,
fontsize=\footnotesize,
linenos
]
{python}
#input batch*4*4 or batch*3*3
#output torch batch*3 x, y, z in radiant
#the rotation is in the sequence of x,y,z
def compute_euler_angles_from_rotation_matrices(rotation_matrices):
    batch = rotation_matrices.shape[0]
    R = rotation_matrices
    sy = torch.sqrt(R[:,0,0]*R[:,0,0]+R[:,1,0]*R[:,1,0])
    singular = sy<1e-6
    singular = singular.float()
        
    x = torch.atan2(R[:,2,1], R[:,2,2])
    y = torch.atan2(-R[:,2,0], sy)
    z = torch.atan2(R[:,1,0],R[:,0,0])
    
    xs = torch.atan2(-R[:,1,2], R[:,1,1])
    ys = torch.atan2(-R[:,2,0], sy)
    zs = R[:,1,0]*0
        
    gpu = rotation_matrices.get_device()
    if gpu < 0:
        out_euler = torch.autograd.Variable(torch.zeros(batch,3)).to(torch.device('cpu'))
    else:
        out_euler = torch.autograd.Variable(torch.zeros(batch,3)).to(torch.device('cuda:%d' % gpu))
    out_euler[:,0] = x*(1-singular)+xs*singular
    out_euler[:,1] = y*(1-singular)+ys*singular
    out_euler[:,2] = z*(1-singular)+zs*singular
        
    return out_euler
\end{minted}
\caption{6D-RepNet/6D-RepNet360's Euler angle extraction code}
\label{listing:12}
\end{listing}
\begin{listing}
\begin{minted}
[
frame=single,
framesep=2mm,
baselinestretch=1.2,
%bgcolor=LightGray,
fontsize=\footnotesize,
linenos
]
{python}
def get_R(x,y,z):
    ''' Get rotation matrix from three rotation angles (radians). right-handed.
    Args:
        angles: [3,]. x, y, z angles
    Returns:
        R: [3, 3]. rotation matrix.
    '''
    # x
    Rx = np.array([[1, 0, 0],
                   [0, np.cos(x), -np.sin(x)],
                   [0, np.sin(x), np.cos(x)]])
    # y
    Ry = np.array([[np.cos(y), 0, np.sin(y)],
                   [0, 1, 0],
                   [-np.sin(y), 0, np.cos(y)]])
    # z
    Rz = np.array([[np.cos(z), -np.sin(z), 0],
                   [np.sin(z), np.cos(z), 0],
                   [0, 0, 1]])

    R = Rz.dot(Ry.dot(Rx))
    return R
\end{minted}
\caption{6D-RepNet/6D-RepNet360's rotation matrix formula}
\label{listing:4}
\end{listing}

\begin{listing}[!ht]
\begin{minted}
[
frame=single,
framesep=2mm,
baselinestretch=1.2,
%bgcolor=LightGray,
fontsize=\footnotesize,
linenos
]
{python}
def draw_axis(img, yaw, pitch, roll, tdx=None, tdy=None, size = 100):
    pitch = pitch * np.pi / 180
    yaw = -(yaw * np.pi / 180)
    roll = roll * np.pi / 180

    if tdx != None and tdy != None:
        tdx = tdx
        tdy = tdy
    else:
        height, width = img.shape[:2]
        tdx = width / 2
        tdy = height / 2

    # X-Axis pointing to right. drawn in red
    x1 = size * (cos(yaw) * cos(roll)) + tdx
    y1 = size * (cos(pitch) * sin(roll) + cos(roll) * sin(pitch) * sin(yaw)) + tdy

    # Y-Axis | drawn in green
    #        v
    x2 = size * (-cos(yaw) * sin(roll)) + tdx
    y2 = size * (cos(pitch) * cos(roll) - sin(pitch) * sin(yaw) * sin(roll)) + tdy

    # Z-Axis (out of the screen) drawn in blue
    x3 = size * (sin(yaw)) + tdx
    y3 = size * (-cos(yaw) * sin(pitch)) + tdy

    cv2.line(img, (int(tdx), int(tdy)), (int(x1),int(y1)),(0,0,255),4)
    cv2.line(img, (int(tdx), int(tdy)), (int(x2),int(y2)),(0,255,0),4)
    cv2.line(img, (int(tdx), int(tdy)), (int(x3),int(y3)),(255,0,0),4)

    return img
\end{minted}
\caption{HopeNet/WHENet/6D-RepNet/6D-RepNet360's three-line drawing routine}
\label{listing:5}
\end{listing}

\begin{listing}[ht]
\begin{minted}
[
frame=single,
framesep=2mm,
baselinestretch=1.2,
%bgcolor=LightGray,
fontsize=\footnotesize,
linenos
]
{python}
def plot_pose_cube(img, yaw, pitch, roll, tdx=None, tdy=None, size=150.):
    # Input is a cv2 image
    # pose_params: (pitch, yaw, roll, tdx, tdy)
    # Where (tdx, tdy) is the translation of the face.
    # For pose we have [pitch yaw roll tdx tdy tdz scale_factor]

    p = pitch * np.pi / 180
    y = -(yaw * np.pi / 180)
    r = roll * np.pi / 180
    if tdx != None and tdy != None:
        face_x = tdx - 0.50 * size 
        face_y = tdy - 0.50 * size

    else:
        height, width = img.shape[:2]
        face_x = width / 2 - 0.5 * size
        face_y = height / 2 - 0.5 * size

    x1 = size * (cos(y) * cos(r)) + face_x
    y1 = size * (cos(p) * sin(r) + cos(r) * sin(p) * sin(y)) + face_y 
    x2 = size * (-cos(y) * sin(r)) + face_x
    y2 = size * (cos(p) * cos(r) - sin(p) * sin(y) * sin(r)) + face_y
    x3 = size * (sin(y)) + face_x
    y3 = size * (-cos(y) * sin(p)) + face_y

    # Draw base in red
    cv2.line(img, (int(face_x), int(face_y)), (int(x1),int(y1)),(0,0,255),3)
    cv2.line(img, (int(face_x), int(face_y)), (int(x2),int(y2)),(0,0,255),3)
    cv2.line(img, (int(x2), int(y2)), (int(x2+x1-face_x),int(y2+y1-face_y)),(0,0,255),3)
    cv2.line(img, (int(x1), int(y1)), (int(x1+x2-face_x),int(y1+y2-face_y)),(0,0,255),3)
    # Draw pillars in blue
    cv2.line(img, (int(face_x), int(face_y)), (int(x3),int(y3)),(255,0,0),2)
    cv2.line(img, (int(x1), int(y1)), (int(x1+x3-face_x),int(y1+y3-face_y)),(255,0,0),2)
    cv2.line(img, (int(x2), int(y2)), (int(x2+x3-face_x),int(y2+y3-face_y)),(255,0,0),2)
    cv2.line(img, (int(x2+x1-face_x),int(y2+y1-face_y)), 
                   (int(x3+x1+x2-2*face_x),int(y3+y2+y1-2*face_y)),(255,0,0),2)
    # Draw top in green
    cv2.line(img, (int(x3+x1-face_x),int(y3+y1-face_y)), (int(x3+x1+x2-2*face_x),
                   int(y3+y2+y1-2*face_y)),(0,255,0),2)
    cv2.line(img, (int(x2+x3-face_x),int(y2+y3-face_y)), 
                   (int(x3+x1+x2-2*face_x),int(y3+y2+y1-2*face_y)),(0,255,0),2)
    cv2.line(img, (int(x3), int(y3)), (int(x3+x1-face_x),int(y3+y1-face_y)),(0,255,0),2)
    cv2.line(img, (int(x3), int(y3)), (int(x3+x2-face_x),int(y3+y2-face_y)),(0,255,0),2)

    return img
\end{minted}
\caption{6D-RepNet/6D-RepNet360's pose-box drawing routine for the Euler angles}
\label{listing:6}
\end{listing}

\begin{listing}[ht]
\begin{minted}
[
frame=single,
framesep=2mm,
baselinestretch=1.2,
%bgcolor=LightGray,
fontsize=\scriptsize,  
linenos
]
{python}
def save_img_head(frame, save_path, seq, cam, cam_id, json_file, frame_id, threshold, yaw_ref):
    img_path = os.path.join(save_path, seq)
    frame = cv2.cvtColor(frame, cv2.COLOR_BGR2RGB)
    frame = Image.fromarray(frame)
    # print(frame.size)
    E_ref = np.mat([[1, 0, 0, 0.],
                    [0, -1, 0, 0],
                    [0, 0, -1, 50],
                    [0, 0, 0,  1]])
    cam['K'] = np.mat(cam['K'])
    cam['distCoef'] = np.array(cam['distCoef'])
    cam['R'] = np.mat(cam['R'])
    cam['t'] = np.array(cam['t']).reshape((3, 1))
    with open(json_file) as dfile:
        fframe = json.load(dfile)
        count_face = -1
        yaw_avg = 0
    for face in fframe['people']:
        # 3D Face has 70 3D joints, stored as an array [x1,y1,z1,x2,y2,z2,...]
        face3d = np.array(face['face70']['landmarks']).reshape((-1, 3)).transpose()
        face_conf = np.asarray(face['face70']['averageScore'])
        model_points_3D = np.ones((4, 58), dtype=np.float32)
        model_points_3D[0:3] = model_points
        clean_match = (face_conf[kp_idx] > 0.1) #only pick points confidence higher than 0.1
        kp_idx_clean = kp_idx[clean_match]
        kp_idx_model_clean = kp_idx_model[clean_match]
        if(len(kp_idx_clean)>6):
            count_face += 1
            rotation, translation, error, scale = align(np.mat(model_points_3D[0:3, kp_idx_model_clean]),
                                                        np.mat(face3d[:, kp_idx_clean]))
            sphere_new = scale * rotation @ (sphere) + translation
            pt_helmet = projectPoints(sphere_new,
                                               cam['K'], cam['R'], cam['t'],
                                               cam['distCoef'])
            temp = np.zeros((4, 4))
            temp[0:3, 0:3] = rotation
            temp[0:3, 3:4] = translation
            temp[3, 3] = 1
            E_virt = np.linalg.inv(temp @ np.linalg.inv(E_ref))
            E_real = np.zeros((4, 4))
            E_real[0:3, 0:3] = cam['R']
            E_real[0:3, 3:4] = cam['t']
            E_real[3, 3] = 1

            compound = E_real @ np.linalg.inv(E_virt)
            status, [pitch, yaw, roll] = select_euler(np.rad2deg(inverse_rotate_zyx(compound)))
            yaw= -yaw
            roll = -roll
            yaw_avg = yaw_avg+yaw
            if(abs(yaw-yaw_ref)>threshold or yaw_ref==-999):
                if (status == True):
                    # record the yaw, pitch, roll to the annotation file
                        :
\end{minted}
\caption{WHENet's head pose extraction code}
\label{listing:11}
\end{listing}

\begin{listing}[ht]
\begin{minted}
[
frame=single,
framesep=2mm,
baselinestretch=1.2,
%bgcolor=LightGray,
fontsize=\footnotesize,
linenos
]
{python}
def inverse_rotate_zyx(M):
    if np.linalg.norm(M[:3, :3].T @ M[:3, :3] - np.eye(3)) > 1e-5:    
        raise ValueError('Matrix is not a rotation')

    if np.abs(M[0, 2]) > 0.9999999:
        # gimbal lock
        z = 0.0
        # M[1,0] =  cz*sx*sy
        # M[2,0] =  cx*cz*sy
        if M[0, 2] > 0:   
            y = -np.pi / 2   
            x = np.arctan2(-M[1, 0], -M[2, 0])  # Note that M11 = -M20
        else:                 
            y = np.pi / 2    
            x = np.arctan2(M[1, 0], M[2, 0])
        return np.array((x, y, z)), np.array((x, y, z))
    else:
        # no gimbal lock
        y0 = np.arcsin(-M[0, 2])
        y1 = np.pi - y0
        cy0 = np.cos(y0)
        cy1 = np.cos(y1)

        x0 = np.arctan2(M[1, 2] / cy0, M[2, 2] / cy0)
        x1 = np.arctan2(M[1, 2] / cy1, M[2, 2] / cy1)

        z0 = np.arctan2(M[0, 1] / cy0, M[0, 0] / cy0)
        z1 = np.arctan2(M[0, 1] / cy1, M[0, 0] / cy1)
        return np.array((x0, y0, z0)), np.array((x1, y1, z1))

def select_euler(two_sets):
    pitch, yaw,  roll= two_sets[0]
    pitch2, yaw2, roll2 = two_sets[1]
    if yaw>180.:
        yaw = yaw - 360.
    if yaw2>180.:
        yaw2 = yaw2 - 360.
    if abs(roll)<90 and abs(pitch)<90:
        return True, [pitch, yaw, roll]
    elif abs(roll2)<90 and abs(pitch2)<90:
        return True, [pitch2, yaw2, roll2]
    else:
        return False, [-999, -999, -999]

\end{minted}
\caption{WHENet's Euler angle extraction}
\label{listing:0}
\end{listing}

\begin{listing}[ht]
\begin{minted}
[
frame=single,
framesep=2mm,
baselinestretch=1.2,
%bgcolor=LightGray,
fontsize=\scriptsize,
linenos
]
{python}
def auto_labels_generating(imgPath, filter_joints_list):

    valid_bbox_euler_list = []

    lost_faces = 0
    for [face2d, occlusion, face3d, camR, camT, camK] in filter_joints_list: 
        # face3d has 51 3D joints, stored as an array with shape [51,3]
        face3d = np.array(face3d).reshape((-1, 3)).transpose()

        rotation, translation, error, scale = align_3d_head(
            np.mat(model_points_3D[0:3, kp_idx_model]), np.mat(face3d[:, kp_idx_agora]))
        
        sphere_new = scale * rotation @ (sphere) + translation
        pt_helmet = projectPoints(sphere_new, camK, camR, camT, [0,0,0,0,0])
            
        temp = np.zeros((4, 4))
        temp[0:3, 0:3] = rotation
        temp[0:3, 3:4] = translation
        temp[3, 3] = 1
        E_virt = np.linalg.inv(temp @ np.linalg.inv(E_ref))
        
        E_real = np.zeros((4, 4))
        E_real[0:3, 0:3] = camR
        E_real[0:3, 3:4] = camT
        E_real[3, 3] = 1

        compound = E_real @ np.linalg.inv(E_virt)
        status, [pitch, yaw, roll] = select_euler(np.rad2deg(inverse_rotate_zyx(compound)))
        yaw = -yaw
        roll = -roll

        if status == True:
            x_min = int(max(min(pt_helmet[0, :]),0))
            y_min = int(max(min(pt_helmet[1, :]),0))
            x_max = int(min(max(pt_helmet[0, :]), img_w))
            y_max = int(min(max(pt_helmet[1, :]), img_h))
            w, h = x_max-x_min, y_max-y_min
            '''
            Exclude heads with "out-of-bounding position", "severely truncated" or "super-large size".
            However, we still could not filter out heads "without face labels" or "totally occluded".
            '''
            # sanity check
            if x_min<x_max and y_min<y_max and h/w<1.5 and w/h<1.5 and w<img_w*0.7 and h<img_h*0.7:  
                head_bbox = [x_min, y_min, w, h]  # format [x,y,w,h]
                euler_angles = [pitch, yaw, roll]  # represented by degree
                valid_bbox_euler_list.append({
                    "face2d_pts": [list(face2d[:,0]), list(face2d[:,1])], 
                    "head_bbox": head_bbox, 
                    "euler_angles": euler_angles})
            else:
                # print("The face in this frame is not having valid face bounding box...")
                continue
        else:
            # print("The face in this frame is not having valid three Euler angles...")
            lost_faces += 1
            continue

    return valid_bbox_euler_list, lost_faces
\end{minted}
\caption{DirectMHP's auto label generation}
\label{listing:7}
\end{listing}

\begin{listing}[ht]
\begin{minted}
[
frame=single,
framesep=2mm,
baselinestretch=1.2,
%bgcolor=LightGray,
fontsize=\footnotesize,
linenos
]
{python}
def reference_head(scale=0.01,pyr=(10.,0.0,0.0)):
    kps = np.asarray([[-7.308957, 0.913869, 0.000000], [-6.775290, -0.730814, -0.012799],
        [-5.665918, -3.286078, 1.022951], [-5.011779, -4.876396, 1.047961],
        [-4.056931, -5.947019, 1.636229], [-1.833492, -7.056977, 4.061275],
        [0.000000, -7.415691, 4.070434], [1.833492, -7.056977, 4.061275],
        [4.056931, -5.947019, 1.636229], [5.011779, -4.876396, 1.047961],
        [5.665918, -3.286078, 1.022951],
        [6.775290, -0.730814, -0.012799], [7.308957, 0.913869, 0.000000],
        [5.311432, 5.485328, 3.987654], [4.461908, 6.189018, 5.594410],
        [3.550622, 6.185143, 5.712299], [2.542231, 5.862829, 4.687939],
        [1.789930, 5.393625, 4.413414], [2.693583, 5.018237, 5.072837],
        [3.530191, 4.981603, 4.937805], [4.490323, 5.186498, 4.694397],
        [-5.311432, 5.485328, 3.987654], [-4.461908, 6.189018, 5.594410],
        [-3.550622, 6.185143, 5.712299], [-2.542231, 5.862829, 4.687939],
        [-1.789930, 5.393625, 4.413414], [-2.693583, 5.018237, 5.072837],
        [-3.530191, 4.981603, 4.937805], [-4.490323, 5.186498, 4.694397],
        [1.330353, 7.122144, 6.903745], [2.533424, 7.878085, 7.451034],
        [4.861131, 7.878672, 6.601275], [6.137002, 7.271266, 5.200823],
        [6.825897, 6.760612, 4.402142], [-1.330353, 7.122144, 6.903745],
        [-2.533424, 7.878085, 7.451034], [-4.861131, 7.878672, 6.601275],
        [-6.137002, 7.271266, 5.200823], [-6.825897, 6.760612, 4.402142],
        [-2.774015, -2.080775, 5.048531], [-0.509714, -1.571179, 6.566167],
        [0.000000, -1.646444, 6.704956], [0.509714, -1.571179, 6.566167],
        [2.774015, -2.080775, 5.048531], [0.589441, -2.958597, 6.109526],
        [0.000000, -3.116408, 6.097667], [-0.589441, -2.958597, 6.109526],
        [-0.981972, 4.554081, 6.301271], [-0.973987, 1.916389, 7.654050],
        [-2.005628, 1.409845, 6.165652], [-1.930245, 0.424351, 5.914376],
        [-0.746313, 0.348381, 6.263227], [0.000000, 0.000000, 6.763430],
        [0.746313, 0.348381, 6.263227], [1.930245, 0.424351, 5.914376],
        [2.005628, 1.409845, 6.165652], [0.973987, 1.916389, 7.654050],
        [0.981972, 4.554081, 6.301271]]).T  # 58 3D points
    R = rotate_zyx( np.deg2rad(pyr) )
    kps = transform( R, kps*scale )
    tris = Delaunay( kps[:2].T ).simplices.copy()
    return kps, tris
\end{minted}
\caption{WHENet's auto label generation}
\label{listing:8}
\end{listing}

\end{appendices}
\end{document}